%% file: arxiv.tex
\documentclass[lettersize,journal]{IEEEtran}
\usepackage{amsmath,amsfonts}
\usepackage{algorithmic}
\usepackage[ruled,vlined,linesnumbered]{algorithm2e}
\usepackage{array}
\usepackage[caption=false,font=normalsize,labelfont=sf,textfont=sf]{subfig}
\usepackage{textcomp}
\usepackage{stfloats}
\usepackage{url}
\usepackage{verbatim}
\usepackage{graphicx}
\def\BibTeX{{\rm B\kern-.05em{\sc i\kern-.025em b}\kern-.08em
    T\kern-.1667em\lower.7ex\hbox{E}\kern-.125emX}}
\usepackage{balance}
\usepackage{paralist}
\usepackage{marvosym}
\usepackage{float}
\usepackage{booktabs}
\usepackage{hyperref}
\input{macro}

\usepackage{capt-of}
\usepackage{color}
\usepackage[HTML]{xcolor}
\usepackage{colortbl}
\usepackage{tabularx} 
\usepackage{booktabs}
\usepackage{multirow}    
\usepackage{makecell}
\definecolor{mycolor_blue}{HTML}{E7EFFA}
\definecolor{mycolor_bluelight}{HTML}{E6E6FA}
\definecolor{mycolor_green}{HTML}{E6F8E0}
\definecolor{mycolor_gray}{HTML}{ECECEC}
\definecolor{mycolor_yellow}{HTML}{FFF8DC}
\definecolor{pearDark}{HTML}{2980B9}
\definecolor{color2}{RGB}{230,247,224}
\definecolor{color1}{HTML}{C7EAFE}

\definecolor{cvprblue}{rgb}{0.21,0.49,0.74}

\begin{document}
\title{EmoAgent: A Multi-Agent Framework for Diverse Affective Image Manipulation}
\author{Qi Mao$^{1,\textsuperscript{\Letter}}$ \hspace{0.65cm} Haobo Hu$^{1}$ \hspace{0.65cm} Yujie He$^{1}$ \hspace{0.65cm} Difei Gao$^2$ \hspace{0.65cm} Haokun Chen$^{1}$ \hspace{0.65cm} Libiao Jin$^{1}$\\
$^1$MIPG, Communication University of China \hspace{0.65cm} $^2$Show Lab, National University of Singapore 

\thanks{Qi Mao and Libiao Jin are with the School of Information and Communication Engineering and the State Key Laboratory of Media Convergence and Communication, Communication University of China, Beijing 100024, China (E-mail: $\{$qimao, libiao$\}$@cuc.edu.cn). \\
Haobo Hu, Yujie He and Haokun Chen are with the School of Information and Communication Engineering, Communication University of China, Beijing 100024, China (E-mail: $\{$hhaobo, hyj$\}$@mails.cuc.edu.cn, chenhaokun@cuc.edu.cn).\\
Difei Gao is with Show Lab, National University of Singapore (E-mail: daniel.difei.gao@gmail.com).
}
}

\markboth{Journal of \LaTeX\ Class Files,~Vol.~14, No.~8, August~2021}%
{Shell \MakeLowercase{\textit{et al.}}: A Sample Article Using IEEEtran.cls for IEEE Journals}


\maketitle
\let\thefootnote\relax\footnotetext{
\textsuperscript{\Letter} Corresponding Author}

\input{sec/0_abstract}
\input{sec/1_intro}

\input{sec/2_related_work}

\input{sec/3_method}
\input{sec/4_experiments}

\input{sec/5_Discussion}
\input{sec/6_Conclusions}

\IEEEpubidadjcol
\bibliographystyle{IEEEtran}
\bibliography{main}

\newpage
\appendix
\input{sec/X_suppl}

\vfill
\end{document}

%% file: macro.tex
\usepackage{xspace}
\usepackage{color}
\usepackage{multirow}
\usepackage{ifthen}

\renewcommand{\baselinestretch}{1}

\long\def\ignorethis#1{}

\definecolor{gray}{rgb}{0.35,0.35,0.35}
\definecolor{MyBlue}{rgb}{0,0.2,0.8}
\definecolor{MyRed}{rgb}{0.8,0.2,0}
\definecolor{MyGreen}{rgb}{0.0,0.5,0.1}
\definecolor{MyGray}{rgb}{0.4,0.4,0.4}

\newlength\paramargin
\newlength\figmargin
\newlength\subfigmargin
\newlength\secmargin
\newlength\subsecmargin
\newlength\tabmargin
\newlength\eqmargin

\usepackage{array}
\newcolumntype{L}[1]{>{\raggedright\let\newline\\\arraybackslash\hspace{0pt}}m{#1}}
\newcolumntype{C}[1]{>{\centering\let\newline\\\arraybackslash\hspace{0pt}}m{#1}}
\newcolumntype{R}[1]{>{\raggedleft\let\newline\\\arraybackslash\hspace{0pt}}m{#1}}

\def\ie{i.e.,~}
\def\eg{e.g.,~}

\def\vs{vs.~}

\newcommand{\secref}[1]{Section~\ref{sec:#1}}

\newcommand{\figref}[1]{Fig.~\ref{fig:#1}}
\newcommand{\tabref}[1]{Table~\ref{tab:#1}}

\newcommand{\Paragraph}[1]{\noindent\textbf{#1}}

%% file: sec/0_abstract.tex
\begin{abstract}
Affective Image Manipulation (AIM) aims to alter visual elements within an image to evoke specific emotional responses from viewers.
However, existing AIM approaches rely on rigid \emph{one-to-one} mappings between emotions and visual cues, making them ill-suited for the inherently subjective and diverse ways in which humans perceive and express emotion.
To address this, we introduce a novel task setting termed \emph{Diverse AIM (D-AIM)}, aiming to generate multiple visually distinct yet emotionally consistent image edits from a single source image and target emotion.
We propose \emph{EmoAgent}, the first multi-agent framework tailored specifically for D-AIM. 
EmoAgent explicitly decomposes the manipulation process into three specialized phases executed by collaborative agents: a Planning Agent that generates diverse emotional editing strategies, an Editing Agent that precisely executes these strategies, and a Critic Agent that iteratively refines the results to ensure emotional accuracy. 
This collaborative design empowers EmoAgent to model \emph{one-to-many} emotion-to-visual mappings, enabling semantically diverse and emotionally faithful edits.
Extensive quantitative and qualitative evaluations demonstrate that EmoAgent substantially outperforms state-of-the-art approaches in both emotional fidelity and semantic diversity, effectively generating multiple distinct visual edits that convey the same target emotion.
\end{abstract}


%% file: sec/1_intro.tex
\section{Introduction}
\begin{figure*}[!t]
    \centering
    \includegraphics[width=\linewidth]{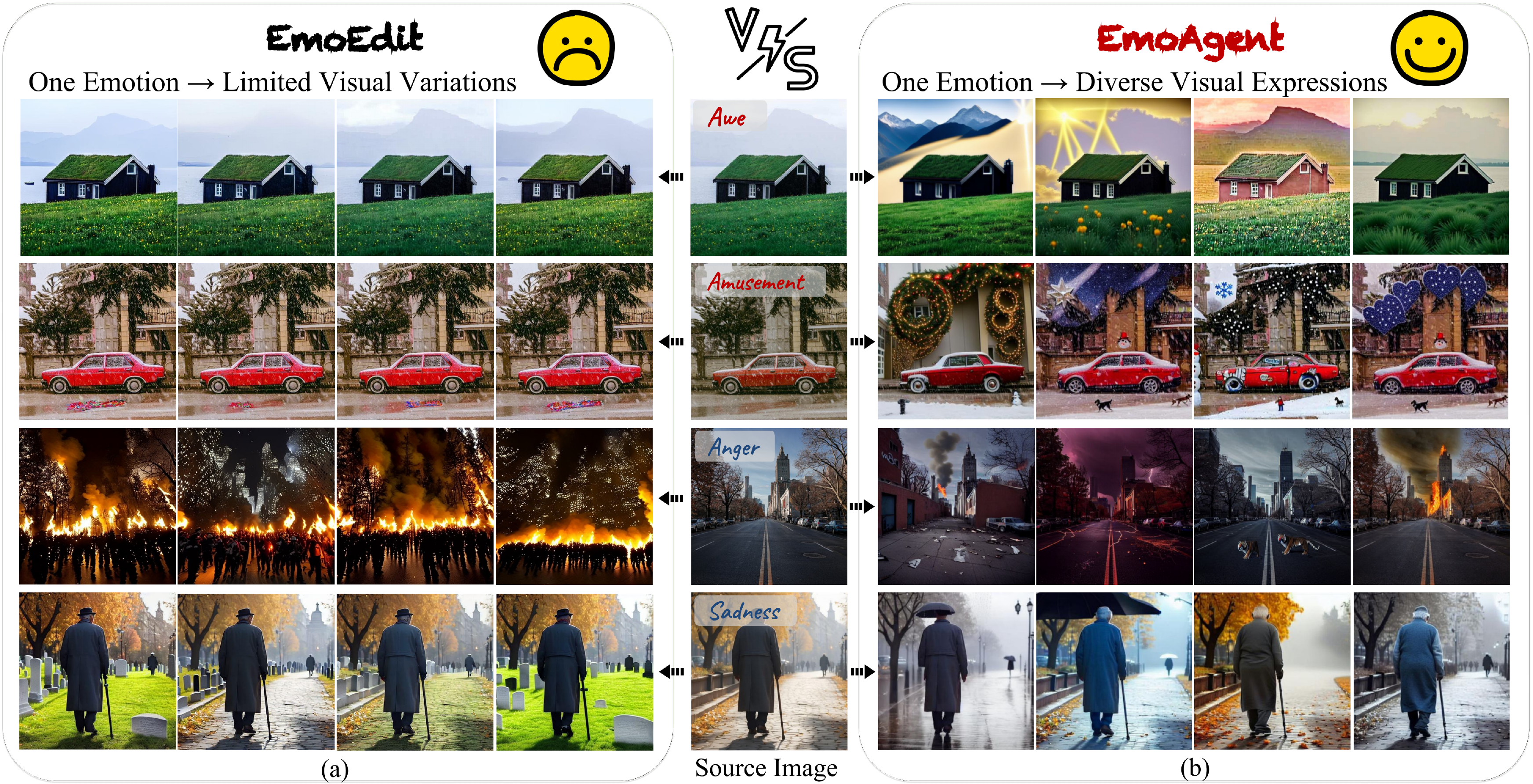}
    \vspace{-7mm}
    \caption{
       \textbf{EmoEdit~\cite{yang2024emoedit} \vs EmoAgent on the D-AIM task.}
(a) Existing approaches typically follow a fixed one-to-one mapping from emotion to visual output, leading to limited diversity.
(b) In contrast, EmoAgent performs multi-path planning and editing to produce multiple semantically distinct yet emotionally consistent results for the same input and target emotion.
    }
    \label{fig:teaser}
\vspace{-7mm}
\end{figure*}

\IEEEPARstart{I}{n} our everyday lives, images serve as powerful conveyors of emotion—a family photo that captures the warmth of a reunion, a travel snapshot that brings back the excitement of adventure, or an advertisement designed to evoke desire or happiness. 
Manipulating these images to enhance or alter their emotional impact is not just about filters and adjustments; it's about harnessing visual elements to deepen emotional impact and create a meaningful connection with viewers.
For instance, a marketer might want to adjust an image to make a product appear more appealing, or a social media user might edit a photo to better reflect the mood of a memorable moment. 
This leads to the goal of \emph{affective image manipulation (AIM)}: modifying images to evoke specific emotional responses in viewers.

Early AIM work focuses on mapping emotions to low-level adjustments~\cite{weng2023affective,fu2022language}, such as modifying color tones or styles, which fall short of conveying meaningful emotional content.
Later methods encode target emotion labels as global feature vectors~\cite{lin2024make}, but fail to capture the fine-grained variations inherent in affective expression.
The state-of-the-art (SOTA) EmoEdit~\cite{yang2024emoedit} augments a semantic editor with a plug-in Emotional Adapter to enable affective content-aware editing, yet still associates each target emotion with a single visual pattern.
As shown in \figref{teaser}(a), this results in a \emph{one-to-one mapping between emotion and visual expression, overlooking the inherently subjective and diverse nature of affective perception.}

Emotions, however, are often interpreted through varied visual cues.
As demonstrated in \figref{teaser}(b), the emotion ``awe'' may be evoked by snow-capped mountain peaks, a dramatic stage spotlight, a field of daisies, or a vivid coastal sunset.
``Amusement'' may appear through colorful toys, playful poses, or warm ambient scenes,
while ``sadness'' can emerge from rainy streets, umbrellas, or grey-toned stillness.
Despite their visual diversity, these scenes convey the same emotional intent.
This underscores a key challenge for AIM:
\emph{How can models generate emotionally consistent yet visually diverse edits that reflect the subjective variability in emotional perception?}

To address this gap, we extend the standard AIM paradigm and introduce
\emph{\textbf{Diverse-AIM (D-AIM)}}, a task that aims to generate multiple visually distinct yet emotionally consistent images from the same source image and target emotion.
We cast D-AIM as a multi-solution space-exploration problem, viewing AIM as the search for diverse yet emotion-consistent transformations. %
Building on this formulation, we present \textbf{\emph{EmoAgent}}, a novel multi-agent framework for D-AIM in which a team of collaborative agents explores
different semantic editing paths to produce varied and emotionally faithful
results.

Specifically, our proposed multi-agent framework achieves emotionally rich and diverse image edits by assigning specialized roles to three collaborative agents, as illustrated in \figref{pipeline}(a). 
During the pre-creation stage, the Planning Agent, equipped with a vision-language model (VLM) and an emotion-factor knowledge (EFK) retriever, analyses the input image, retrieves emotion-relevant cues, and synthesizes diverse editing plans that span a wide range of emotion-relevant semantic elements.
These plans are then passed to the Editing Agent, which translates them into a sequence of editing operations, injecting the specified visual elements into the source image with step-wise precision.

However, the reliability of AIM is not fully assured by these two agents alone. 
A third component, the Critic Agent, acting as the ``eyes'', evaluates intermediate results and collaborates with the Editing Agent in an iterative loop to refine outputs during the optimization stage, ensuring that the target emotion is conveyed both accurately and coherently.
Through this collaborative design, EmoAgent not only enhances emotional alignment but also supports emotion-aware visual diversity—capturing multiple valid interpretations of the same affective goal, as demonstrated in \figref{teaser}(b).

Our contributions are summarized as follows:
\begin{itemize}

\item We introduce \textbf{Diverse AIM (D-AIM)}, a new task setting within AIM that focuses on generating multiple, visually distinct yet emotionally consistent outputs from a single image and target emotion.

\item We propose \textbf{EmoAgent}, the first multi-agent framework for D-AIM, integrating Planning, Editing, and Critic agents to enable diverse visual expressions under the same emotional intent.

\item We conduct extensive quantitative and qualitative evaluations, demonstrating that EmoAgent significantly outperforms prior methods in both emotional accuracy and visual diversity within the same emotion.
\end{itemize}

%% file: sec/2_related_work.tex
\section{related works}

\subsection{Affective Computing}
Affective computing~\cite{picard2000affective} focuses on enabling machines to perceive and interpret human emotions through various modalities such as images and text. 
In visual emotion understanding, two primary emotion representation paradigms are commonly adopted: Categorical Emotional States (CES), which assign discrete emotion labels (e.g., Ekman's six basic emotions~\cite{ekman1992argument}, Mikels' eight-class model~\cite{mikels2005emotional}), and Dimensional Emotional Spaces (DES), which represent emotions continuously along dimensions such as valence and arousal.

Early studies on image emotion analysis rely on handcrafted features and conventional machine learning models, which lacked the capacity to capture the nuanced interplay between image semantics and affective signals. With the rise of deep learning, emotion understanding has significantly improved, thanks to multi-stage perception pipelines~\cite{10464196}, attention mechanisms~\cite{9397355}, and cross-modal techniques such as emotion-aware generation~\cite{10380727} and semantic-affective fusion.

Recently, large pre-trained language and vision-language models have further advanced the field. Text-based models like BERT~\cite{devlin2018bert} and RoBERTa~\cite{liu2019roberta} exhibit strong capabilities in emotion recognition from text, while multimodal models such as GPT-4V~\cite{brown2020language,10572294} can reason about emotions in visual and textual contexts. These advancements open new opportunities for integrating affective computing into generative content creation, especially in emotionally aligned tasks.

In this work, we incorporate multimodal large language models (MLLMs) into the AIM task, enabling the system to analyze and evaluate emotional consistency during editing. 
This integration is essential for achieving semantically coherent and emotionally expressive image edits.

\begin{figure*}[!t]
    \centering
    \includegraphics[width=\linewidth]{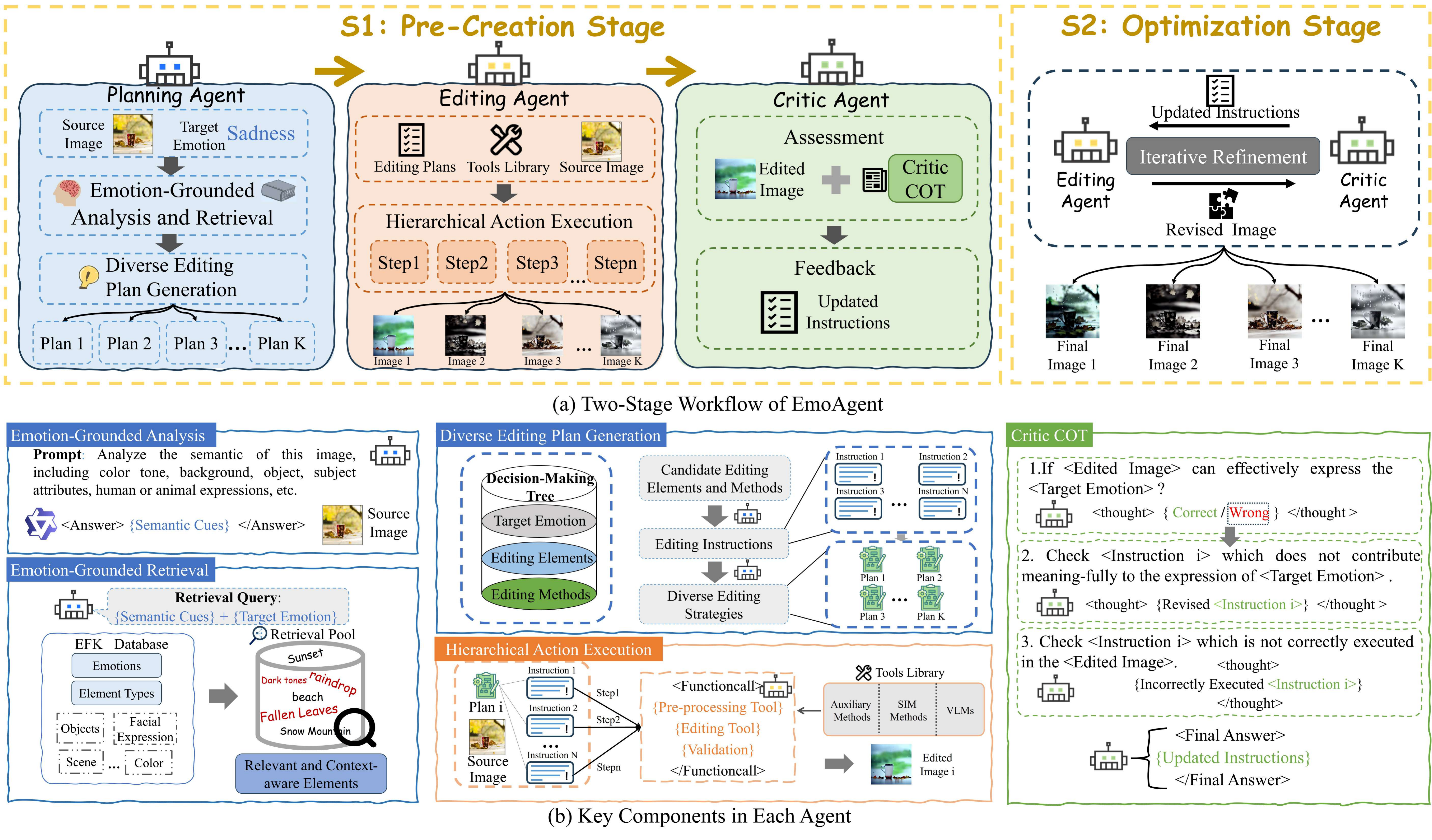} 
    \vspace{-8 mm}
        \caption{
\textbf{Performing Diverse-AIM using EmoAgent.} 
(a) Overview of the EmoAgent framework, which operates in two stages: \textit{pre-creation} and \textit{optimization}. In the pre-creation stage, the Planning Agent generates multiple editing plans, which the Editing Agent executes to produce diverse affective outputs. The Critic Agent evaluates each result’s emotional accuracy; unsatisfactory cases are passed to the optimization stage, where the Editing and Critic Agents iteratively refine the outputs.
(b) Key modules within each agent: emotion-grounded analysis-retrieval and diverse editing plan generation (Planning Agent); hierarchical action execution (Editing Agent); and chain-of-thought (CoT)-based emotional critique (Critic Agent). 
 Together, these modules enable EmoAgent to generate emotionally consistent yet visually diverse edits from the same source image and target emotion.}
      \label{fig:pipeline}
        \vspace{-6mm}
\end{figure*}

\vspace{-3mm}
\subsection{Affective Image Manipulation}
Traditional semantic image manipulation (SIM) approaches perform image edits by interpreting and applying the explicit semantic instructions contained in text prompts.
Built upon diffusion models~\cite{ho2020denoising,dhariwal2021diffusion,rombach2022high}, recent SIM methods typically employ cross-attention tuning~\cite{hertz2022prompt}, instruction-based editing~\cite{brooks2023instructpix2pix}, or mask-guided operations~\cite{avrahami2023blended,avrahami2022blended}.  
While effective for semantic transformations, they are generally not designed to interpret or express emotional semantics (\eg ``make it exciting''), which limits their applicability in tasks that require affective reasoning and emotionally grounded edits.

To address this limitation, recent work has explored AIM, which focuses on modifying images to evoke targeted emotional responses from the viewers, rather than simply aligning with textual descriptions. 
Early approaches in this domain focus on integrating emotion-conditioned image generation, where models synthesize new images aligned with emotional goals.
For instance, EmoGen~\cite{yang2024emogen} maps discrete emotion categories to visual attributes, while Affective-Conditioned Image Generation~\cite{10541050} employs a multi-layer perceptron to adjust generated content across three affective dimensions.

Subsequent works have shifted toward manipulating existing images to convey target emotions.
Early efforts in this direction mainly rely on low-level adjustments, such as modifying lighting or style to indirectly evoke emotional effects~\cite{weng2023affective,fu2022language}, which often fall short in capturing complex emotional nuances without manipulating semantic conetent elements.
Later works, such as Make Me Happier~\cite{lin2024make}, incorporate emotional cues into the editing process through global feature vectors or feature-level embeddings.
EmoEdit~\cite{yang2024emoedit} further advances this line of work by integrating an Emotion Adapter module into diffusion-based editing pipelines, enabling end-to-end emotional control within SIM architectures.

Despite recent advances, existing AIM methods only rely on a one-to-one mapping between emotions and visual features, where each emotion is represented by a fixed set of elements or styles for a given image.
However, emotional expression is inherently diverse—multiple distinct modifications can evoke the same emotion depending on scene semantics, visual composition, and narrative intent.
As such, we extend the AIM into \textbf{D-AIM}, which aims to generate multiple semantically distinct yet emotionally consistent image edits for a given input and target emotion.

\vspace{-1mm}
\subsection{LLM-As-Agent}
\vspace{-1mm}
LLM-based agents have gained traction due to their strong reasoning and decision-making capabilities~\cite{song2023llm,singh2023progprompt,zhou2022least,yao2022react,wei2022chain}. 
Approaches like Chain of Thought~\cite{wei2022chain} and ReAct~\cite{yao2022react} enhance LLMs’ ability to handle complex planning and sequential reasoning through prompt-based workflows.
%
More recently, these agent-driven paradigms have been extended to the visual domain~\cite{wu2023visual,yang2023mm,gao2024assistgui,schumann2024velma}, enabling more interactive and adaptive solutions for vision-language understanding tasks.

In the context of visual generation, GenArtist~\cite{wang2024genartist} introduces a \emph{single-agent} framework that decomposes textual prompts into sequential generation and editing steps for SIM.
However, AIM demands deeper emotional reasoning and the ability to translate affective intent into semantically grounded visual edits—often requiring multi-step planning and refinement beyond literal semantic alignment.

%

%
To meet these demands, we propose \textbf{EmoAgent}, a \emph{multi-agent} framework composed of a Planning Agent, Editing Agent, and Critic Agent. 
In contrast to single-agent systems that often lack explicit planning and critic capabilities, our architecture supports emotion-aware reasoning and multi-step decision-making, enabling collaborative and emotionally aligned editing results.

%% file: sec/3_method.tex
\section{Multi-Agent Framework for Diverse-AIM}
\label{sec:Method}

In this section, we present our approach to addressing the D-AIM task, where the objective is to generate multiple visually distinct image edits from a single source image and target emotion, while ensuring emotional consistency across all outputs.
To address this, we propose \textbf{EmoAgent}, a novel multi-agent framework designed to decompose the editing process into specialized roles. 
As illustrated in \figref{pipeline}(a), EmoAgent comprises three collaborative agents:
\emph{Planning}, \emph{Editing}, and \emph{Critic}—which work together to achieve emotionally rich and visually diverse image generation.
Each agent contributes to a key function:
\begin{compactitem}
      \item the \textbf{Planning Agent} translates emotional intent into diverse editing strategies.
    \item the \textbf{Editing Agent} executes these strategies via controlled image modifications.
    \item  the \textbf{Critic Agent} evaluates emotional correctness and provides feedback. 
\end{compactitem}

This division of labor supports a closed-loop editing pipeline that progressively refines results to align with both emotional and visual goals.
We begin by formalizing the D-AIM task in \secref{overview}, followed by an overview of the multi-agent collaboration workflow in \secref{workflow}. The remaining subsections then detail the mechanisms employed by each agent in \secref{plan}, \secref{edit}, and \secref{critic}.

\vspace{-3mm}
\subsection{Task Definition: Diverse-AIM}
\label{sec:overview}
In this work, we introduce a new task setting within AIM, termed \textbf{Diverse-AIM (D-AIM)}, which aims to generate multiple visually distinct versions of a given image that all express the same target emotion category $E_t$.
Following EmoEdit~\cite{yang2024emoedit}, we adopt Mikels’ eight-class emotion categories~\cite{mikels2005emotional} as implemented in the EmoSet dataset~\cite{yang2023emoset}.
Formally, given a source image $I_o$ and a target emotion label $E_t$, the objective is to produce a set of edited images $\{I_t^{(1)}, I_t^{(2)}, \dots, I_t^{(k)}\}$ such that:
(i) each $I_t^{(k)}$ is visually and semantically distinct from the others;
(ii) the high-level semantic structure and visual layout of the source image $I_o$ are preserved;
(iii) all edited images accurately express the intended emotion $E_t$.

\vspace{-3mm}
\subsection{Agent Collaboration Workflow}
\label{sec:workflow}
To ensure both emotional fidelity and visual diversity, EmoAgent operates in a two-stage workflow: a Pre-Creation Stage for generating diverse candidates, and an Optimization Stage for refining outputs that fail to convey the intended emotion. 
This process is illustrated in \figref{pipeline}(a).

1) \textbf{Pre-Creation Stage.}
Given a source image $I_o$ and a target emotion $E_t$, the Planning Agent first generates a set of diverse editing plans $\mathcal{P} = \{P^{(1)}, P^{(2)}, \dots, P^{(k)}\}$, where each plan represents a distinct strategy for conveying the target emotion. 
The Editing Agent then executes each plan, translating the corresponding editing plans to produce a set of preliminary results $\{I_t^{(1)}, I_t^{(2)}, \dots, I_t^{(k)}\}$. 
Each result $I_t^{(k)}$ is independently evaluated by the Critic Agent to determine whether it effectively conveys the target emotion.
If the emotion is successfully conveyed, $I_t^{(k)}$ is accepted as a valid output; otherwise, it is passed to the Optimization Stage for further refinement.

2) \textbf{Optimization Stage.} 
In this phase, each preliminary result $I_t^{(k)}$ that fails to meet the emotional criteria enters an independent feedback loop.
The Critic Agent identifies necessary revisions to improve emotional alignment, and the Editing Agent applies the corresponding edits.
This iterative refinement continues for each unsatisfactory image until a final result $I_t^{(k)}$ is produced that satisfies the target emotion.
All refined outputs ${I_t^{(k)}}$ together form the final diverse emotional edits.

\begin{figure}[!t]
    \centering
    \includegraphics[width=\columnwidth]{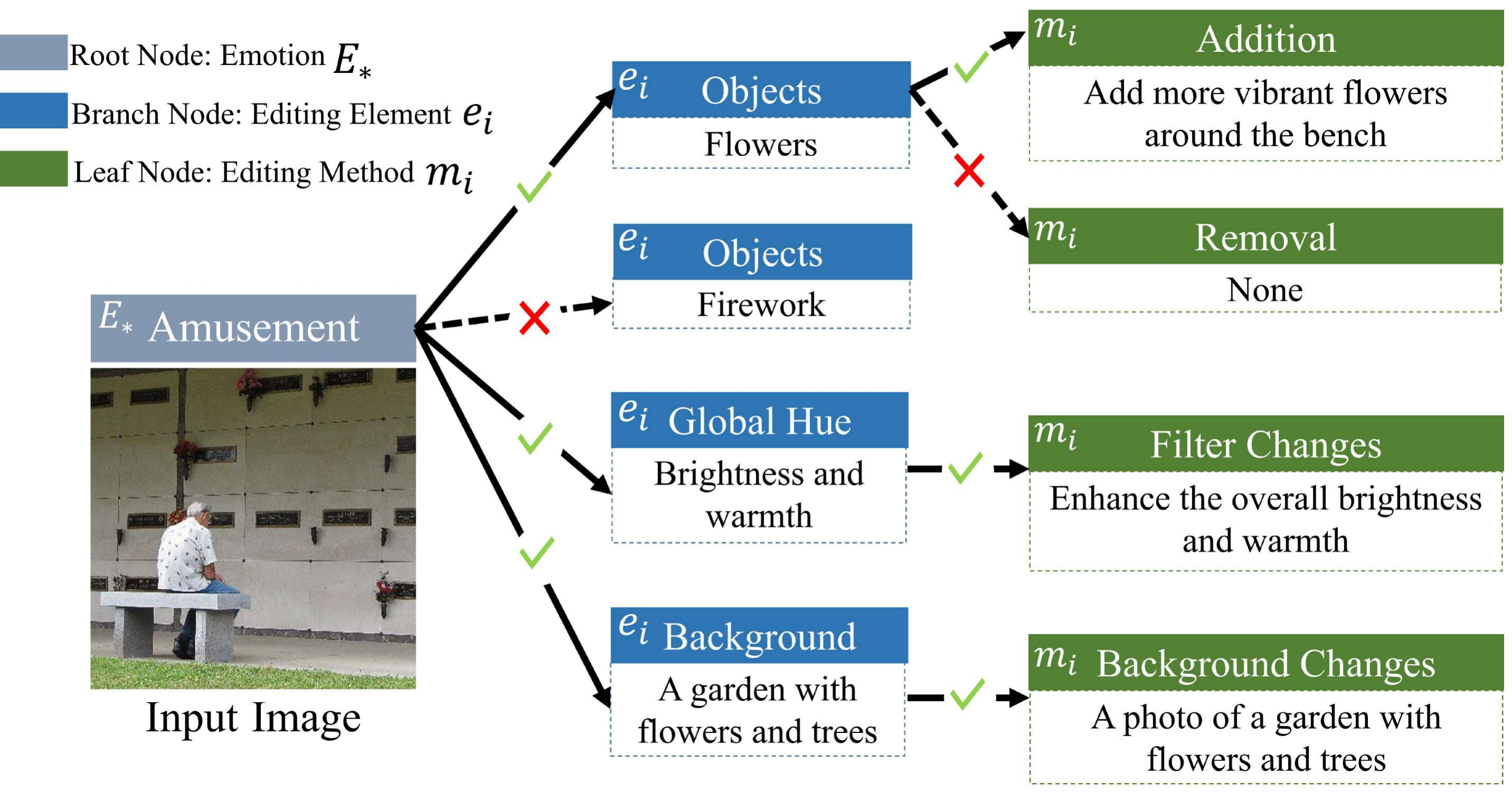} 
    \caption{
        \textbf{Three-layer decision space}.
       The agent designates the target emotion as the root node, identifies appropriate editing elements as branch nodes, and assigns corresponding editing methods as leaf nodes. 
    }
    \vspace{-6mm}
    \label{fig:plan_space}
\end{figure}

\vspace{-6mm}
\subsection{Planning Agent \includegraphics[scale=0.09,bb=0 15 400 34]{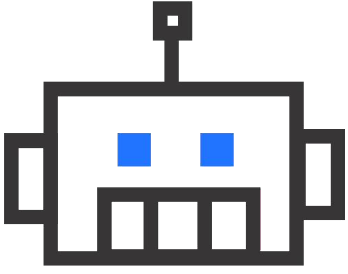}}
\vspace{-1 mm}
\label{sec:plan}
At the heart of EmoAgent is the Planning Agent, which reformulates the D-AIM task as a \emph{multi-solution generation problem}.
Instead of producing a single optimal edit, it aims to map the same target emotion $E_t$ to \emph{multiple semantically distinct yet emotionally consistent editing strategies}.
This diversity is achieved by transforming emotional cues into structured editing instructions via a combination of analysis, retrieval, and planning.

\Paragraph{Input-Output Schema.}
The Planning Agent takes as input a source image $I_o$ and a target emotion $E_t$, and outputs a set of diverse editing plans $\mathcal{P} = \{P^{(1)}, P^{(2)}, \dots, P^{(k)}\}$.
Each $P^{(k)} = \{\text{Ins}^{(k)}_1, \text{Ins}^{(k)}_2, \dots\}$ defines a coherent sequence of editing instructions, representing a distinct transformation path.
Together, these plans offer multiple visually diverse solutions to express the same emotional intent.

\Paragraph{Emotion-Grounded Analysis and Retrieval.}
The Planning Agent begins by using a VLM (\eg Qwen2.5-VL~\cite{Qwen2.5-VL}) to extract semantic cues $S_o$ and the source emotion $E_o$ from the input image $I_o$.
These cues guide a Retrieval-Augmented Generation (RAG)~\cite{lewis2020retrieval} process to query an external Emotion-Factor Knowledge (EFK) database, constructed from EmoSet~\cite{yang2023emoset} and organized into emotion–element–method triples, as illustrated in \figref{pipeline}(b).
Unlike prior approaches that retrieve a single best match, we extract a diverse top-$k$ pool of editing elements that are both semantically relevant to $S_o$ and emotionally aligned with the target emotion $E_t$.
The agent further prioritizes elements most aligned with the visual semantics of $I_o$, ensuring that each editing path remains faithful to the context of the original image.
This retrieved pool serves as a foundation for constructing diverse editing strategies, each capturing a distinct visual interpretation of $E_t$.

\begin{figure}[!t]
    \centering
    \includegraphics[width=\columnwidth]{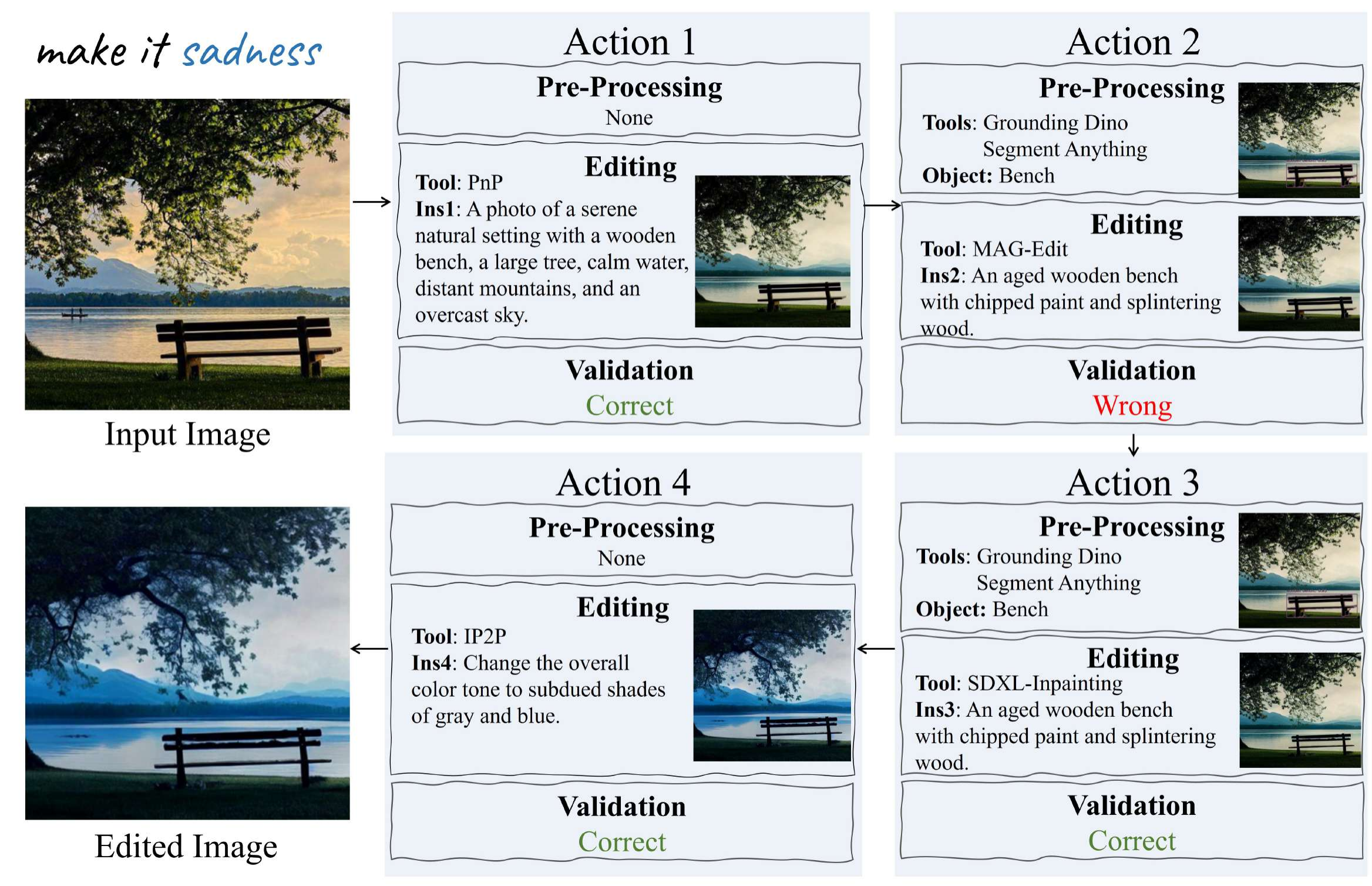} 
    \vspace{-7mm}
    \caption{
     \textbf{Trajectory of actions selected in the action space}.
        If an action fails to execute successfully, a new action is generated and re-executed to ensure the desired outcome.
    }
\label{fig:action_node}
\vspace{-5mm}
\end{figure}

\Paragraph{Diverse Editing Plan Generation.}
To convert retrieved editing elements into structured plans, the Planning Agent explores a three-layer decision space, as shown in \figref{plan_space}:
\begin{itemize}
\item \textbf{Root node:}
The emotion types $E_{\ast}$, which initiate the editing plan and encompass eight emotional labels from EmoSet~\cite{yang2023emoset}: amusement, awe, contentment, excitement, anger, disgust, fear, and sadness. 
\item \textbf{Branch node:} The editing elements $e_i$, including various components such as object semantics, 
attributes, facial expressions, backgrounds, and global hue.
\item \textbf{Leaf node:} The editing methods $m_i$, consisting of manipulation operations such as object replacement, addition, removal, expression changes, filter changes, background changes, and attribute changes.
\end{itemize}

Guided by the target emotion $E_t$ and semantic cues $S_o$, the Planning Agent first selects a set of emotionally relevant and context-aware elements $e_i$ from the retrieved pool, and then pairs each with an appropriate manipulation method $m_i$.
 These pairs $(e_i, m_i)$ are converted into natural-language instructions via a formatting function $f$, which indicates the editing intent (\eg, ``replace the background with a sunset'').
Each instruction is denoted as:
\begin{equation} 
{\rm{Ins}}_i = f(e_i, m_i).
\end{equation}

By assembling different instruction sets, the agent constructs a set of $k$ candidate editing plans:
\begin{align}
\mathcal{P} &= \{ P^{(k)} \mid k = 1, \dots, K \}, \\
P^{(k)} &= \{ \text{Ins}_n^{(k)} \mid n = 1, \dots, N \},
\end{align}
Each plan $P^{(k)}$ represents an independent visual manipulation path, semantically coherent with $I_o$ and emotionally aligned with $E_t$.
This design enables the agent to move beyond fixed mappings and \emph{generate diverse, interpretable editing strategies that reflect distinct expressions of the same target emotion.}

\subsection{Editing Agent \includegraphics[scale=0.09,bb=0 15 400 34]{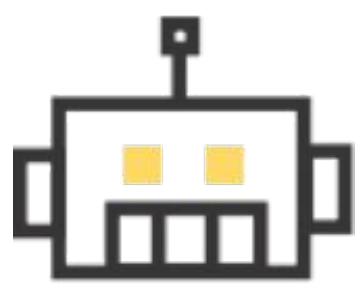}}
\label{sec:edit}
\vspace{-1 mm}

The Editing Agent serves as the executor within the EmoAgent framework—analogous to the "hands" of an artist—responsible for translating abstract editing plans into concrete visual modifications. 
It leverages a comprehensive library of specialized editing tools, functioning as a diverse set of brushes, to faithfully implement each step of the editing process with precision and emotional alignment.

\Paragraph{Input-Output Schema.}
The Editing Agent takes the editing plan $\mathcal{P} = \{P^{(1)}, P^{(2)}, \dots, P^{(k)}\}$ and the source image $I_o$ as input, producing emotionally consistent yet visually diverse edited images $\{I_t^{(1)}, I_t^{(2)}, \dots, I_t^{(k)}\}$ aligned with the target emotion $E_t$.

\Paragraph{Editing Tools Library.}
To support diverse visual transformations, the Editing Agent is equipped with an extensible library of tools categorized by input modality, as shown in \tabref{example}: (1) text-guided tools, (2) tools combining text and spatial masks, and (3) tools integrating spatial masks with reference images.
Each category offers unique strengths for specific editing tasks, such as object insertion, background replacement, and fine-grained expression or attribute manipulation.
To support precise, localized edits, we further incorporate auxiliary tools, including object detection~\cite{liu2023grounding} and segmentation models~\cite{kirillov2023segment}, which provide spatial priors and instance-level understanding.
In addition, a VLM~\cite{Qwen2.5-VL} is employed as a self-critic module to validate the semantic fidelity of the outputs.
Consequently, the Editing Agent dynamically selects the most appropriate tool based on the semantics and complexity of each instruction, ensuring accurate and context-aware execution.

\begin{table}[!t]
  \caption{\textbf{Tools library in the Editing Agent.}\label{tab:example}}
  \vspace{-3mm}
  \renewcommand\arraystretch{1.5}
  \centering
  \resizebox{\linewidth}{!}{%
    \begin{tabular}{ccc}
      \toprule[1pt]
      \textbf{Tools Type} & \textbf{Editing Tools of $\rm{Act}^i_{\rm{edit}}$} & \textbf{Editing Methods}\\
      \midrule
      
      \multirow{5}{*}{\centering{Text-guided}} 
        & Magicbrush~\cite{zhang2024magicbrush} & Objects Addition\\
        & Plug-and-Play~\cite{tumanyan2023plug} & Background Editing\\
        & Guide~\cite{titov2024guide} & Expression Editing\\
        & Ip2p~\cite{brooks2023instructpix2pix} & Filter Editing\\
        & RF-Solver-Edit~\cite{wang2024taming} & \makecell{Object Attribute Editing/\\ Objects Replacement}\\
      \midrule
      
      \multirow{2}{*}{\centering{Text and Masks-based} }
        & SDXL-Inpainting~\cite{podell2023sdxl} & Objects Replacement\\
        & MAG-Edit~\cite{mao2024mag} & Object Attribute Editing\\ 
      \midrule
      
      \centering{\makecell{Masks with Reference\\ Images-based} } 
        & Mimicbrush~\cite{chen2024mimicbrush} & \makecell{Objects Addition/\\ Objects Replacement}\\
      
      \midrule[1pt]
      
      \textbf{Auxiliary Function} & \textbf{Auxiliary Tools} & \textbf{Action Step}\\
      \midrule
      Detection & Grounding Dino~\cite{liu2023grounding} & $\rm{Act}^i_{\rm{pre}}$\\
      Segmentation & Segment Anything~\cite{kirillov2023segment} & $\rm{Act}^i_{\rm{pre}}$\\
      Self-Critic & Qwen2.5-VL~\cite{Qwen2.5-VL} & $\rm{Act}^i_{\rm{val}}$\\
      
      \bottomrule
    \end{tabular}
  }
  \vspace{-3mm}
\end{table}

\Paragraph{Hierarchical Action Execution.}
To implement each editing plan $P^{(k)}$, the Editing Agent executes a sequence of hierarchical actions that progressively refine the image. 
As illustrated in \figref{action_node}, each instruction $\text{Ins}_n^{(k)}$ is handled through three structured stages:
\begin{compactitem}
\item \textbf{Pre-processing ($\rm{Act}^n_{\rm{pre}}$)}: prepares the image for editing by invoking localization models  (\eg object detection~\cite{liu2023grounding}, and segmentation~\cite{kirillov2023segment}) to extract spatial priors when needed.
\item \textbf{Editing ($\rm{Act}^n_{\rm{edit}}$)}: carries out the visual transformation using the appropriate editing tool, selected based on the semantics and modality of  $\text{Ins}_n^{(k)}$.

\item \textbf{Validation ($\rm{Act}^n_{\rm{val}}$)}: evaluates the semantic consistency and emotional alignment of the edited result through the validation actions of the self-critic tool~\cite{Qwen2.5-VL}.
If the validation fails, the instruction is re-executed or adjusted accordingly.
\end{compactitem}
This structured pipeline ensures that the Editing Agent performs accurate, emotionally faithful, and flexible manipulations with diverse and semantically aligned edits.

\vspace{-4 mm}
\subsection{Critic Agent \includegraphics[scale=0.09,bb=0 15 400 34]{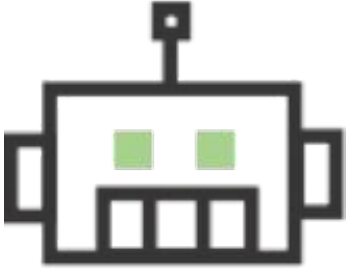}}
\vspace{-1 mm}
\label{sec:critic}
To ensure that emotional intent is faithfully conveyed through visual edits, the Critic Agent functions as the ``eyes'' of the EmoAgent—mirroring the human cognitive cycle of \emph{observation}, \emph{reflection}, and \emph{revision}.
Its primary role is to verify and optimize the editing outcomes by identifying mismatches between the intended emotion $E_t$ and the emotion perceived from the edited image $I_t^{(k)}$.
Operating across both the Pre-Creation and Optimization stages, the Critic Agent leverages Chain-of-Thought (CoT) prompting~\cite{wei2022chain} to perform structured reasoning and deliver precise emotional evaluations.
This enables fine-grained feedback and effective coordination with the Editing Agent to ensure emotional accuracy and reliable execution.

\begin{figure*}[!t]
    \centering
    \includegraphics[width=\textwidth]{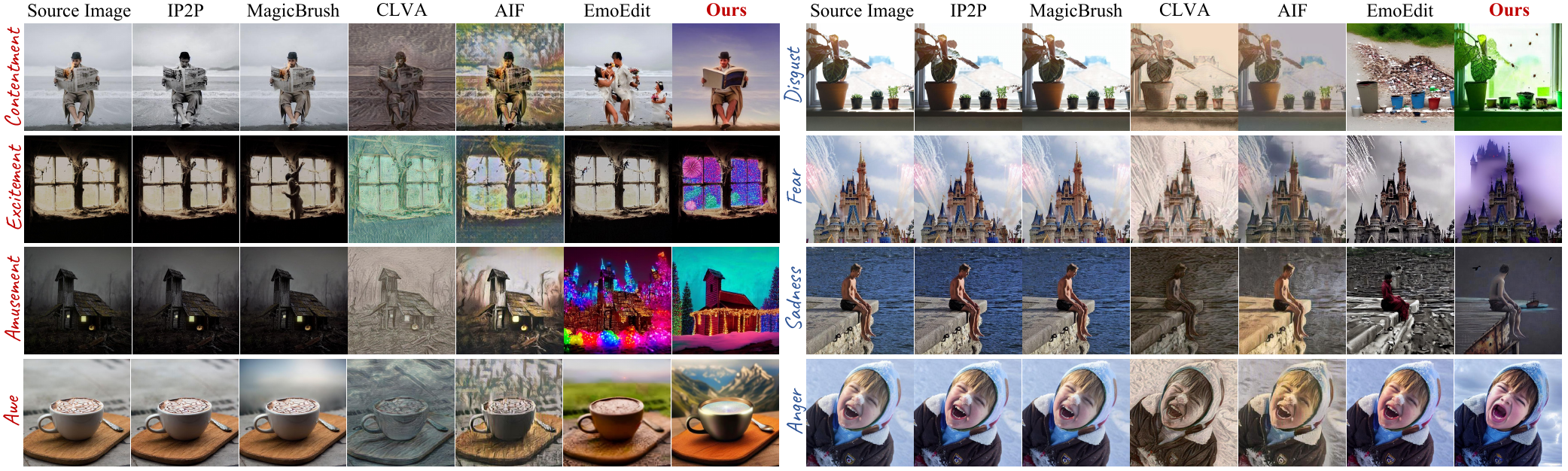} 
    \vspace{-7mm}
        \caption{
            \textbf{Qualitative comparisons with the SIM and AIM baselines.}
EmoAgent more effectively conveys target emotions while preserving high-level semantic content, outperforming prior SIM- and AIM-based methods.
        }
        \label{fig:baseline_1}
    \vspace{-5mm}
\end{figure*}

\Paragraph{Assessment and Feedback.}
Upon receiving an initial edited result $I_t^{(k,0)}$, the Critic Agent first determines whether it successfully conveys the target emotion $E_t$.
If so, the image is finalized as $I_t^{(k,\text{final})}$ and returned as the output.
If not, the agent performs a two-level diagnostic on each editing instruction $\text{Ins}_i$:
\begin{compactitem}
\item \textbf{Emotional Effectiveness:} Does $\text{Ins}_i$ contribute meaningfully to the expression of $E_t$? If not, it is revised into an improved form $\text{Ins}_i^\ast$.
\item \textbf{Execution Accuracy:} If emotionally valid, is $\text{Ins}_i$ correctly executed in the image? If not, the error is flagged for correction.
\end{compactitem}

\Paragraph{Iterative Refinement.}
If any instruction is found to be emotionally ineffective or incorrectly executed, it is revised and returned to the Editing Agent for re-application.
The updated instructions are used to generate a revised image $I_t^{(k, i)}$, which is then re-evaluated by the Critic Agent.
This feedback loop continues until the output image $I_t$ successfully and faithfully conveys the target emotion $E_t$.

This close collaboration between the Critic and Editing Agents enhances the reliability of visual emotional manipulation, ensuring that the final results are both emotionally aligned and semantically faithful.

%% file: sec/4_experiments.tex
\section{EXPERIMENTS}
\label{sec:experiments}

\subsection{Implementation Details}
\Paragraph{Framework Configurations.}
We employ Qwen2.5-VL~\cite{Qwen2.5-VL} as the default VLM and use \texttt{text-embedding-ada-002} 
for text encoding in the EFK Retriever.
All experiments are conducted on
two NVIDIA GeForce RTX $4090$ GPUs.  
All editing and auxiliary tools are implemented using the official released code.
Additional implementation details are
provided in
the supplementary matrial (SM).

\Paragraph{Dataset.} 
To evaluate both the effectiveness and diversity of our emotional editing framework, we use the test set curated in EmoEditSet of EmoEdit~\cite{yang2024emoedit}, which comprises 405 real-world images with varied emotional contexts.  
All input and output images are standardized to a resolution of $512 \times 512$ pixels to ensure fair comparisons across different editing methods.

\Paragraph{Details of EFK Retriever.}
\label{sec:emotree}
We construct the EFK database for EFK retriever by clustering EmoSet~\cite{yang2023emoset} images in CLIP~\cite{radford2021learning} embedding space.
Clusters are refined based on size, pixel-level consistency, and emotional relevance.
We then use GPT-4o~\cite{gpt4o} to generate concise natural language descriptions for each cluster, forming emotion–element pairs.
In practice, the Planning Agent retrieves the top-5 semantically relevant elements from EFK as editing candidates.

\begin{table}[!t]
\caption{\textbf{Quantitative comparisons on Semantic Consistency and Emotion Alignment.} 
\colorbox{color1}{Blue} and \colorbox{color2}{Green} indicate the best and second best value respectively.}
\vspace{-3mm}
\centering
\renewcommand{\arraystretch}{0.8}
\setlength{\fboxsep}{1pt}
\resizebox{\linewidth}{!}{
\begin{tabular}{c|c|ccc}
\toprule
\multicolumn{1}{c}{} & \textbf{Semantic Consistency} & \multicolumn{3}{c}{\textbf{Emotion Alignment}} \\ 
\cmidrule(r){2-2} \cmidrule(r){3-5}
\multicolumn{1}{c}{\multirow{-2}{*}{\textbf{Method/Metrics}}} & \textbf{CLIP-I} \textbf{($\uparrow$)} & \textbf{Emo-A} \textbf{($\uparrow$)} & \textbf{Emo-S} \textbf{($\uparrow$)} & \textbf{ESR} \textbf{($\uparrow$)} \\ 
\midrule
\multicolumn{5}{c}{\textbf{SIM Methods}} \\
\midrule
IP2P~\cite{brooks2023instructpix2pix} & \colorbox{color2}{0.942} & 9.88\% & 0.012 & 54.46\% \\
MagicBrush~\cite{zhang2024magicbrush} & \colorbox{color1}{0.947} & 8.43\% & 0.002 & 50.84\% \\
\midrule
\multicolumn{5}{c}{\textbf{AIM Methods}} \\
\midrule
CLVA~\cite{fu2022language} & 0.765 & 9.88\% & 0.044 & 68.92\% \\
AIF~\cite{weng2023affective} & 0.794 & 7.95\% & 0.031 & 69.40\% \\
EmoEdit~\cite{yang2024emoedit} & 0.797 & \colorbox{color2}{54.46\%} & \colorbox{color2}{0.355} & \colorbox{color2}{89.64\%} \\
\textbf{Ours} & 0.778 & \colorbox{color1}{64.58\%} & \colorbox{color1}{0.357} & \colorbox{color1}{97.59\%} \\
\bottomrule
\end{tabular}
}
\vspace{-5 mm}
\label{tab:OBJECTIVE_METRIC}
\end{table}

\begin{figure*}[!t]
    \centering
    \includegraphics[width=1.\textwidth]{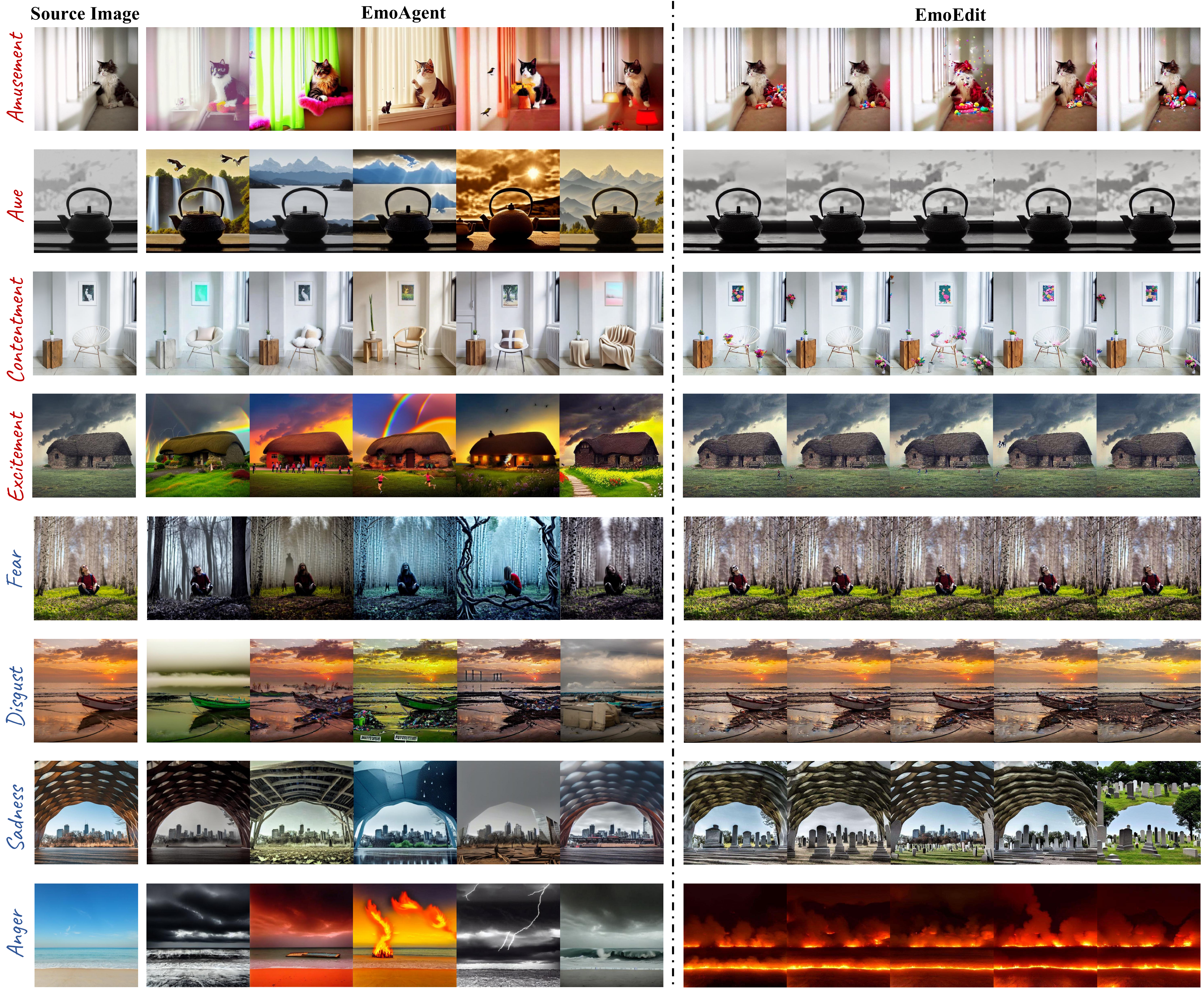} 
    \vspace{-7mm}
    \caption{
\textbf{Emotion-aware visual diversity from a single source image.}
Due to its one-to-one emotion-to-visual mapping, EmoEdit applies fixed editing patterns and struggles to produce diverse outputs. 
In contrast, EmoAgent introduces varied and context-aware visual elements to generate multiple emotionally consistent yet visually distinct edits from the same image.
    }
    \vspace{-7mm}
    \label{fig:diversity1}
\end{figure*}

\subsection{Evaluation Details}
\Paragraph{Evaluation Metrics.}
To demonstrate that EmoAgent can effectively and accurately complete the AIM task while exhibiting more diverse emotional expression capability, we evaluate the performance of our method from three aspects:
\begin{compactitem}
\item \textbf{Semantic Consistency.}
    We evaluate how well the edited images preserve the content of the original using CLIP-I~\cite{radford2021learning} for semantic similarity.

\item \textbf{Emotion Alignment.} 
We evaluate how well the edited image aligns with the intended emotion across three aspects:
Following a similar strategy to~\cite{yang2024emogen}, we assess emotional correctness using Emo-A, which employs EmoVIT~\cite{xie2024emovit} to measure the alignment between edited images and the target emotions.
To further quantify emotional enhancement, we introduce Emo-S, which measures the increase in target emotion intensity from the original image $I_o$ to the edited image $I_t$.
We also include the Emotional Shift Ratio (ESR)~\cite{lin2024make}, which quantifies the overall emotional transformation by computing the Kullback–Leibler Divergence between the emotion distributions of the source and edited images.

\item \Paragraph{Emotion-Aware Diversity.} 
We evaluate the diversity of visual expressions for a given target emotion by measuring perceptual differences across multiple generated outputs. 
Following prior work~\cite{MSGAN,DRIT_plus,mao2022continuous}, we adopt LPIPS as the metric, where higher scores indicate greater visual diversity.
Additionally, we adopt Semantic Diversity (Sem-D)~\cite{yang2024emogen} to evaluate content richness across edited outputs, computed as the average cosine distance between CLIP embeddings of each sample and all other generated variants.
\end{compactitem}

\Paragraph{Evaluation Protocol.}
We begin by evaluating the overall effectiveness of EmoAgent on the AIM task through comparisons in \emph{Semantic Consistency} and \emph{Emotion Alignment} with representative methods from both SIM and AIM domains.
Subsequently, we assess \emph{Emotion-Aware Diversity} by conducting a focused comparison with the SOTA AIM method \ie EmoEdit~\cite{yang2024emoedit}.
For semantic and emotional evaluation, we randomly assign a target emotion to each of the $405$ test images and compute the corresponding metrics.
To evaluate diversity, we randomly select $50$ source images and generate five distinct outputs per image under the same target emotion and experimental settings.
This protocol enables us to measure how effectively EmoAgent generates semantically coherent, emotionally accurate, and visually diverse outputs for a consistent emotional goal.

\begin{figure*}[!t]
    \centering
    \includegraphics[width=\textwidth]{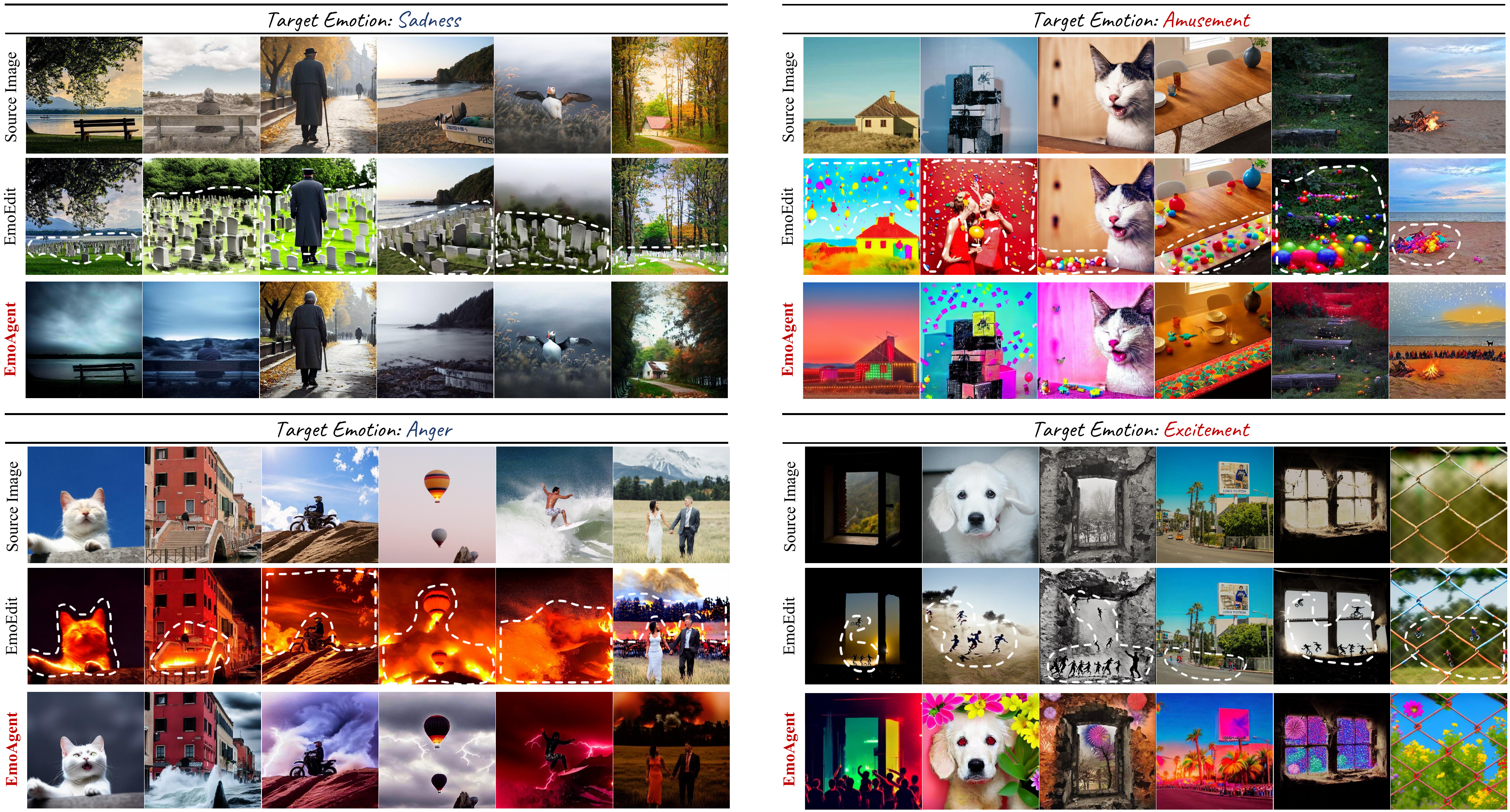} 
    \vspace{-7mm}
    \caption{\textbf{Emotion-aware visual diversity across varied source images.}
Compared to EmoEdit, which tends to reuse fixed emotional elements across different inputs, EmoAgent adapts its edits to each image’s semantics, introducing distinct affective cues and achieving both better emotional fidelity and greater visual diversity.
    }
    \vspace{-1mm}
    \label{fig:diversity2}
\end{figure*}

\subsection{Comparison Results on Semantic Consistency and Emotion Alignment}

\Paragraph{Baselines.}
To evaluate the overall effectiveness of EmoAgent, we compare it with several representative image editing methods in terms of \emph{semantic consistency} and \emph{emotion alignment}.
The selected baselines include both general SIM methods and dedicated AIM approaches:
\begin{itemize}
    \item \textbf{SIM methods:} IP2P~\cite{brooks2023instructpix2pix}, and MagicBrush~\cite{zhang2024magicbrush}, which focus on general-purpose image editing guided by text instructions.
    
    \item \textbf{AIM methods:} Style-based AIM approaches, CLVA~\cite{fu2022language}, AIF~\cite{weng2023affective}, and semantic content-aware AIM, EmoEdit~\cite{yang2024emoedit}, which are specifically designed to manipulate images in alignment with target emotions.   
\end{itemize}

\Paragraph{Qualitative Results.} 
We present qualitative comparisons with baseline methods in \figref{baseline_1}.  
For example, directly applying IP2P~\cite{brooks2023instructpix2pix} yields minimal emotional transformation, with edited images showing little perceptual difference from the original.  
Similarly, style-based AIM methods such as CLVA~\cite{fu2022language} and AIF~\cite{weng2023affective} struggle to convey target emotions effectively through style transfer alone.
Although EmoEdit~\cite{yang2024emoedit} achieves partial emotional transformation, it often suffers from ineffective edits and structural degradation.
For example, in the ``excitement'' and ``anger'' cases, the edits retain original content, such as a ``dilapidated window'' or a ``laughing child'', with minimal changes, failing to enhance emotional expression.
In the ``disgust'' case, EmoEdit replaces ``flower pots'' with ``trash cans'' and modifies the background ``windowsill'' into ``a pile of garbage and grass'', disrupting both semantic fidelity and spatial structure.
In contrast, EmoAgent introduces emotionally relevant cues while preserving image integrity.
For ``excitement'', it adds ``vibrant fireworks'' behind the window; for anger, it enhances tension with ``crying expressions'' and ``lightning''; and for ``disgust'', it subtly incorporates ``dirt'', ``insects'', and a ``greenish hue'', achieving emotional alignment without compromising content or layout.


\Paragraph{Quantitative Results.} 
As illustrated in \tabref{OBJECTIVE_METRIC}, EmoAgent achieves consistently superior performance across all Emotion Alignment metrics, particularly in emotion accuracy.  
Advanced SIM methods such as IP2P~\cite{brooks2023instructpix2pix} and MagicBrush~\cite{zhang2024magicbrush} attain only $9.88$\% and $8.43$\% on the Emo-A metric~\cite{yang2024emogen}, while AIM-specific methods like AIF~\cite{weng2023affective} and CLVA~\cite{fu2022language} perform even worse, ranging from $9.88$\% to $7.95$\%.  
These results highlight the limitations of both SIM and style-based AIM approaches in effectively manipulating affective content.
Moreover, due to the limited capability of understanding affective descriptions, SIM methods such as IP2P~\cite{brooks2023instructpix2pix} and Magicbrush~\cite{zhang2024magicbrush} often produce manipulated images that remain largely unchanged.
This preservation of the original image semantics is a primary reason why these algorithms achieve high scores on the semantic consistency metric.
As a content-aware AIM method, EmoEdit~\cite{yang2024emoedit} goes beyond low-level style modifications by introducing richer semantic content aligned with the target emotion. 
However, due to the lack of an explicit optimization mechanism for emotional alignment, it cannot revise edits that fail to accurately convey the intended emotion.  
Moreover, its reliance on a rigid one-to-one mapping between emotions and visual features limits its capacity to incorporate richer and more context-sensitive affective cues.  
As a result, its performance on emotion-specific metrics—such as Emo-A~\cite{yang2024emogen} and Emo-S—is consistently lower than that of our proposed EmoAgent.
In contrast, EmoAgent achieves the highest scores across all three emotion-related metrics: Emo-A~\cite{yang2024emogen}, ESR~\cite{lin2024make}, and Emo-S, with Emo-A showing a particularly substantial lead over baseline methods.
While the inclusion of more semantic elements may lead to a slight decrease in semantic consistency, EmoAgent still surpasses style-based methods such as CLVA~\cite{fu2022language} in the CLIP-I metric.
These results demonstrate that EmoAgent delivers emotionally faithful and semantically aligned image edits, effectively balancing affective expressiveness and content preservation.

\begin{figure*}[!t]
    \centering
    \includegraphics[width=\textwidth]{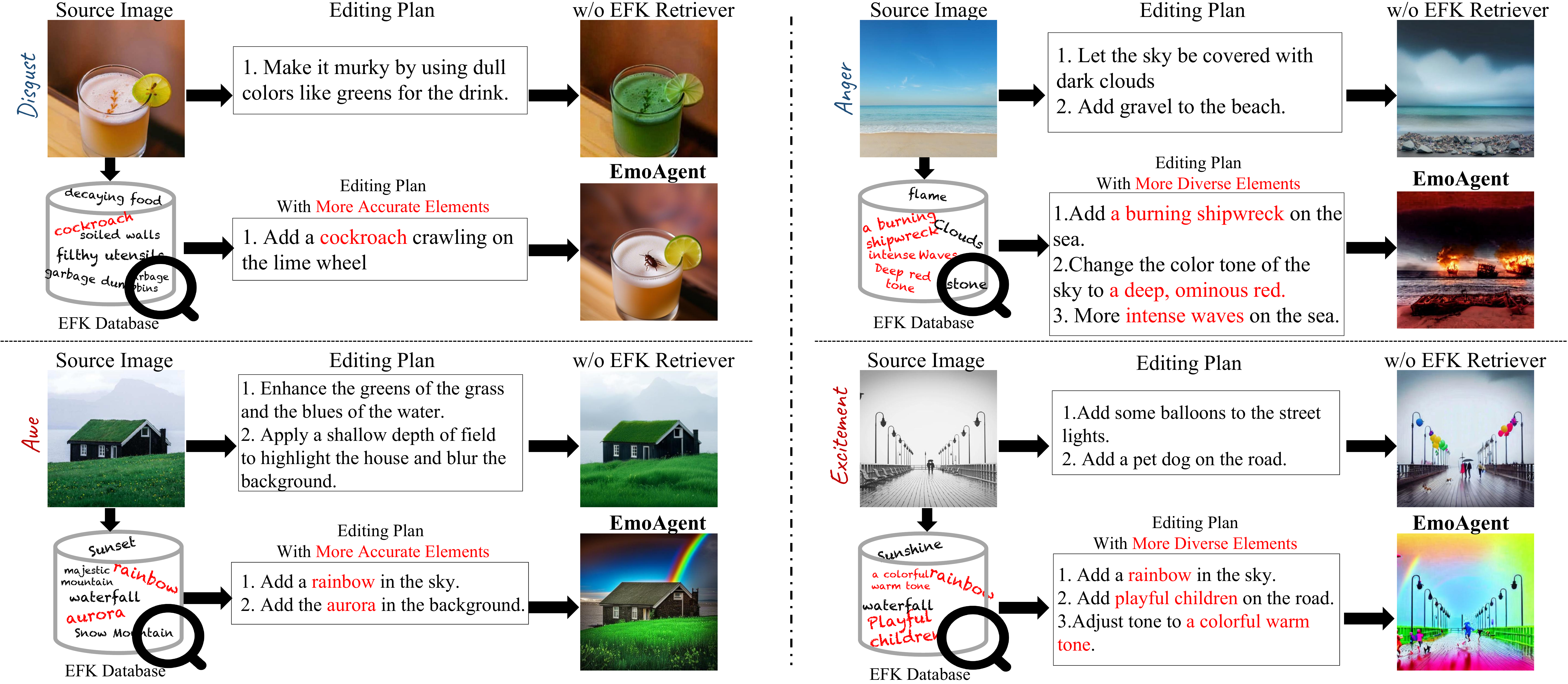} 
    \vspace{-7mm}
    \caption{\textbf{Ablation study on the EFK retriever.}
Removing the EFK retriever reduces both emotional accuracy and diversity. With it, the Planning Agent can retrieve and incorporate a wider range of contextually appropriate emotion-related elements into the editing plan.
    }
    \vspace{-4mm}
    \label{fig:ab_efk}
\end{figure*}

\subsection{Comparison Results on Emotion-Aware Diversity}
To assess the expressive diversity of emotional edits, we compare EmoAgent against the SOTA AIM method EmoEdit~\cite{yang2024emoedit}.

\Paragraph{Qualitative Results.}
We begin by evaluating EmoAgent’s ability to generate \emph{diverse visual expressions} from a single source image under the same target emotion.  
As shown in \figref{diversity1}, EmoEdit~\cite{yang2024emoedit} often relies on repetitive editing elements and operations, frequently reusing the same visual cues across outputs. 
For instance, to express emotions like ``disgust'' or ``amusement,'' it repeatedly inserts foreground elements such as ``garbage dumps'' or ``colorful toys,'' resulting in limited visual variation and constrained emotional representation.
In some cases, such as with ``awe,'' it even fails to effectively convey the intended emotion.
In contrast, EmoAgent, enabled by the collaboration of its three specialized agents, not only ensures accurate emotional alignment, but also generates a broader range of visual features and stylistic variations, all within the same image context.
For example, when expressing ``awe,'' EmoAgent produces visually distinct outputs by flexibly altering both foreground and background elements.
As demonstrated in \figref{diversity1}, the outputs contain a variety of background and atmosphere changes, including ``waterfalls'', ``dazzling sunlight'', ``lakes'', ``vivid purple skies'', and ``distant mountains''. 
The foreground also varies, with additions like ``eagles'', ``delicate textures of pots'', and ``shape of pots'', making each image feel unique while still staying consistent with the original content and emotion.

To further demonstrate EmoAgent’s ability to enhance visual diversity in emotional expression, we extend the comparisons to different source images that share the same target emotion.
For each emotion category, we randomly sample seven input images and examine their edited results.
As shown in \figref{diversity2}, EmoEdit~\cite{yang2024emoedit} often applies the same visual cues across different inputs—such as repeatedly adding ``graves'' for ``sadness'' or ``running people'' for ``excitement''—regardless of the image’s original content.
In contrast, EmoAgent tailors its editing strategy to the semantic context of each image, introducing a wider variety of elements such as ``fireworks,’’ ``tropical tree,’’ and ``flowers,’’ as well as modifying hue, background, facial expressions, and object attributes.
By adapting visual variations to the semantic context of the original image, these edits achieve both emotional accuracy and rich diversity in expression.

\begin{table}[!t]
\caption{\textbf{Quantitative comparisons with EmoEdit on Emotion-Aware Diversity.} \colorbox{color1}{Blue} indicate the best value.}
\centering
\scriptsize
\renewcommand{\arraystretch}{0.8}
\setlength{\fboxsep}{1pt}
\resizebox{0.8\linewidth}{!}{
\begin{tabular}{c|ccc}
\toprule
\multicolumn{1}{c}{} & \multicolumn{2}{c}{\textbf{Emotion-Aware Diversity}}  \\ 
\cmidrule(r){2-3} 
\multicolumn{1}{c}{\multirow{-2}{*}{\textbf{Method/Metrics}}} & \textbf{LPIPS} \textbf{($\uparrow$)}  & \textbf{Sem-D} \textbf{($\uparrow$)}  \\ 
\midrule
EmoEdit~\cite{yang2024emoedit} & 0.205 & 0.086   \\
\textbf{Ours} & \colorbox{color1}{\textbf{0.488}} & \colorbox{color1}{\textbf{0.180}}  \\
\bottomrule
\end{tabular}
}
\vspace{-7 mm}
\label{tab:merged_results}
\end{table}

\Paragraph{Quantitative Results.}
As shown in \tabref{merged_results}, we quantitatively evaluate the emotional visual diversity of EmoAgent compared to EmoEdit~\cite{yang2024emoedit}, using LPIPS and Sem-D as metrics.
EmoAgent achieves a significantly higher LPIPS score of \textbf{0.488} compared to EmoEdit’s $0.205$, indicating greater perceptual variation across generated results.
Similarly, EmoAgent attains a Sem-D score of \textbf{0.180}, more than double that of EmoEdit (0.086), demonstrating its superior ability to introduce diverse semantic content while preserving emotional alignment.
These results confirm that EmoAgent adapts its visual expressions to the input image context, resulting in richer and more varied emotional outputs from the same source image.

\subsection{Ablation Studies}
To assess the impact of each core module in EmoAgent, we systematically disable one of the following components in each experiment: \textit{(i)} the EFK Retriever of the Planning Agent, \textit{(ii)} the editing tools library in the Editing Agent, and \textit{(iii)} the Critic Agent.

\Paragraph{Emotion-Factor Knowledge Retriever.}
The EFK Retriever plays a key role in enhancing both the fidelity and diversity of emotional expression. 
By providing external emotion-grounded visual knowledge, it allows the Planning Agent to identify semantically appropriate and emotionally rich editing elements.
As shown in \figref{ab_efk}, enabling the EFK Retriever significantly enhances EmoAgent’s ability to associate and accurate visual cues with abstract emotions. For example, in the case of ``disgust'', it identifies specific element such as \textit{``cockroach''}, while for ``awe'', it retrieves multiple accurate visuals like \textit{``rainbows''} and \textit{``aurora''}.
Moreover, EmoAgent can provide multiple and  more diverse elements. In the case of  ``excitement'', it enables diverse emotional visual features, including \textit{``burning shipwreck''}, \textit{``deep red sky''}, \textit{``rainbows''}, and \textit{``playful children''}, thereby generating richer and more varied emotional expressions.

Without the retriever, the model struggles to surface relevant semantic concepts, leading to generic or emotionally ambiguous outputs.
This qualitative improvement is also supported by quantitative evidence. 
As shown in \tabref{ablation1} and ~\tabref{ablation2}, removing the EFK Retriever results in a significant drop in both emotion accuracy and diversity, \eg Emo-A declines from \textbf{61\%} to \textbf{40\%}, while Sem-D drops from $0.180$ to $0.161$. 
These results confirm that the EFK Retriever is essential for producing emotionally aligned yet visually diverse outputs, supporting its dual role in grounding expression and enriching visual variation.
\vspace{1mm}
\begin{table}[!t]
\caption{\textbf{Quantitative results of the ablation study on semantic consistency and emotion alignment.} 
\colorbox{color1}{Blue} and \colorbox{color2}{Green} indicate the best and second-best value, respectively.%
We perform ablation studies on a subset of $100$ test images.
}
\centering
\vspace{-3 mm}
\resizebox{\linewidth}{!}{
\begin{tabular}{c|c|ccc}
\toprule 
\multicolumn{1}{c|}{} 
& \textbf{Semantic Consistency} 
& \multicolumn{3}{c}{\textbf{Emotion Alignment}} \\ 
\cmidrule(r){2-2} \cmidrule(r){3-5}
\multicolumn{1}{c|}{\multirow{-2}{*}{\textbf{Method/Metrics}}} 
& \textbf{CLIP-I} ($\uparrow$) 
& \textbf{Emo-A} ($\uparrow$) & \textbf{Emo-S} ($\uparrow$) & \textbf{ESR} ($\uparrow$) \\ 
\midrule
\textit{w/o} EFK retriever  
& \colorbox{color1}{0.813} & \colorbox{color2}{40\%} & 0.151 & 66\%   \\
\textit{w/o} Editing Tool Library 
& \colorbox{color2}{0.808} & 35\% & 0.159 & 75\%  \\
\textit{w/o} Critic Agent
& 0.798 & 28\% & \colorbox{color2}{0.168} & \colorbox{color2}{86\%} \\
\textbf{EmoAgent} 
& 0.763 & \colorbox{color1}{61\%} & \colorbox{color1}{0.318} & \colorbox{color1}{95\%} \\
\bottomrule 
\end{tabular}
}

\label{tab:ablation1}
\end{table}

\begin{figure}[!t]
    \centering
    \includegraphics[width=\linewidth]{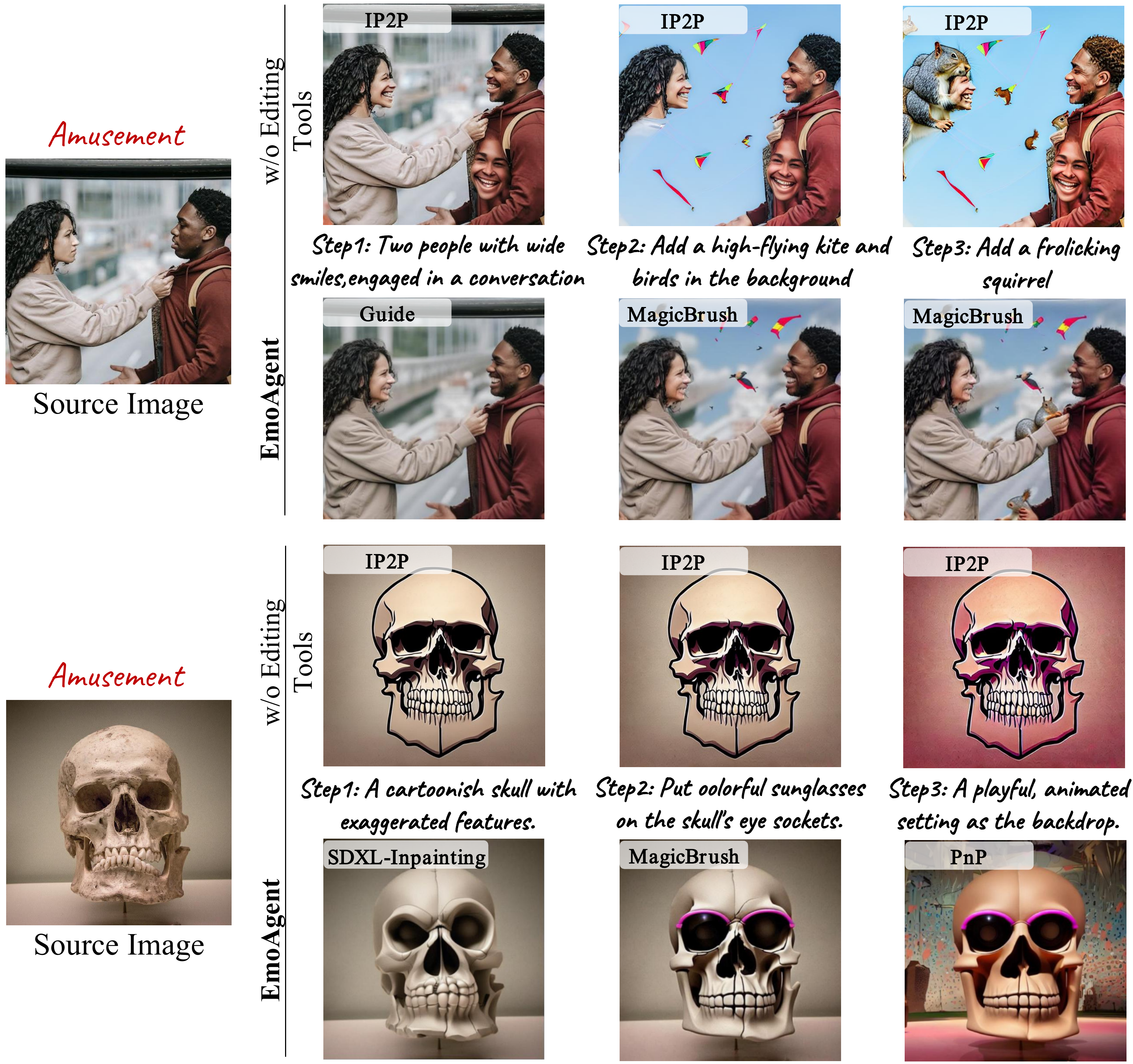} 
    \vspace{-7mm}
    \caption{
            \textbf{Ablation study on the editing tools library.}
Without the full tool library, the system struggles to perform precise or varied visual manipulations. In contrast, EmoAgent dynamically selects appropriate tools based on the editing instructions, enabling more accurate and diverse emotion-aligned edits.
    }
    \vspace{-3mm}
    \label{fig:ab_tool}
\end{figure}

\begin{table}[!t]
\caption{\textbf{Quantitative result of the ablation study on emotion-aware diversity.} 
We assess the impact of the EFK Retriever and Editing Tools on emotion-aware diversity metrics.}
\centering
\renewcommand{\arraystretch}{0.8}
\setlength{\fboxsep}{1pt}
\scriptsize
\resizebox{0.8\linewidth}{!}{
\begin{tabular}{c|cc}
\toprule
\multicolumn{1}{c}{} & \multicolumn{2}{c}{\textbf{Emotion-Aware Diversity}}  \\ 
\cmidrule(r){2-3} 
\multicolumn{1}{c}{\multirow{-2}{*}{\textbf{Method/Metrics}}} & \textbf{LPIPS} \textbf{($\uparrow$)}  & \textbf{Sem-D} \textbf{($\uparrow$)}  \\ 
\midrule
\textit{w/o} EFK Retriver & 0.456 & 0.161  \\
\textit{w/o} Editing Tools & 0.425 & 0.173  \\
\textbf{EmoAgent} & \colorbox{color1}{\textbf{0.488}} & \colorbox{color1}{\textbf{0.180}}  \\
\bottomrule
\end{tabular}
}
\vspace{-5 mm}
\label{tab:ablation2}
\end{table}

\Paragraph{Editing Tools Library.}
The editing tool library is central to EmoAgent’s ability to perform precise and varied visual manipulations. 
By leveraging a diverse set of editing tools—ranging from mask-based to reference-guided methods—the Editing Agent can flexibly handle different types of semantic changes with visual accuracy and emotional relevance.
As shown in ~\figref{ab_tool}, when the system is constrained to using only IP2P~\cite{brooks2023instructpix2pix}, the edits become visibly limited in both quality and granularity. 
For example, in the ``Amusement'' scenario, IP2P alone introduces distortions such as unnatural body shapes, revealing its difficulty in handling localized and attribute-specific edits. 
In contrast, with access to the full toolset, EmoAgent is able to select the most suitable tool for each editing instruction, leading to cleaner, more context-aware transformations—ranging from subtle facial adjustments to large-scale background changes. 
This flexibility not only improves visual plausibility but also contributes to more expressive and diverse emotional interpretations.
These qualitative observations are further supported by quantitative results in ~\tabref{ablation1} and~\tabref{ablation2}. 
Removing the editing tools library leads to a drop in Emo-A from \textbf{61\%} to \textbf{35\%}, and in Emo-S from \textbf{0.318} to \textbf{0.159}, indicating weaker emotional alignment. 
Simultaneously, emotion-aware diversity also suffers: LPIPS decreases from $0.488$ to $0.425$, and Sem-D drops from $0.180$ to $0.173$.
These results confirm that a comprehensive tools library is key to enabling emotionally faithful and visually diverse edits.

\begin{figure}[!t]
    \centering
\includegraphics[width=\linewidth]{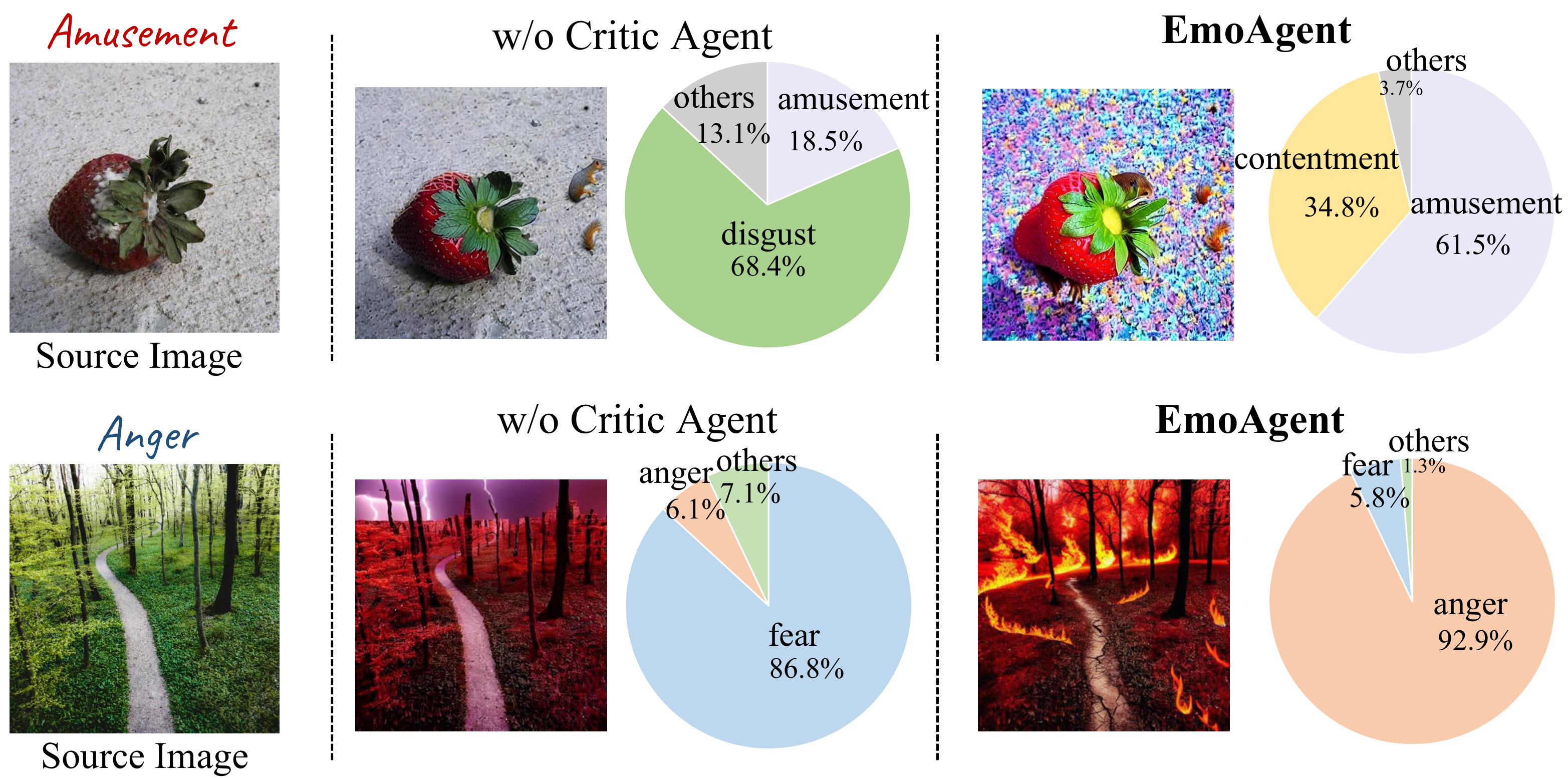} 
    \caption{
        \textbf{Ablation study on the Critic Agent.}
Without the Critic Agent, emotionally inaccurate results are not refined. The Critic Agent enables iterative evaluation and feedback in collaboration with the Editing Agent, improving emotional alignment.
    }
    \label{fig:critic}
   \vspace{-6 mm}
\end{figure}

\Paragraph{Critic Agent.}
The Critic Agent empowers EmoAgent to iteratively refine editing results by identifying and correcting emotionally inaccurate outputs, thereby enhancing expression alignment. 
As shown in \figref{critic}, the ``amusement'' expression is notably improved in the ``strawberry'' example through background and color tone adjustments.  
Removing the Critic Agent leads to the most substantial drop in Emo-A performance (\textbf{61\%} to \textbf{28\%}) among all ablation settings, highlighting its critical role in ensuring emotion accuracy.

\begin{table*}[!tb]
\caption{\textbf{Human Evaluations.} 
Compared to baseline methods, our approach demonstrates superior performance in conveying target emotions, achieving higher favors in emotional accuracy, structural-affective balance, and visual diversity.}
\label{tab:userstudy}
\centering
\vspace{-2 mm}
\resizebox{\linewidth}{!}{\begin{tabular}{cl|cccc}
\toprule
\multicolumn{2}{l|}{\textbf{Methods/Metrics}} 
& \multicolumn{1}{c}{\begin{tabular}[c]{@{}c@{}}\textbf{Emotion Accuracy} ($\uparrow$)\end{tabular}} 
& \multicolumn{1}{c}{\begin{tabular}[c]{@{}c@{}}\textbf{Semantic Plausibility} ($\uparrow$)\end{tabular}} 
& \multicolumn{1}{c}{\begin{tabular}[c]{@{}c@{}}\textbf{Structural-Affective}\\ \textbf{Balance} ($\uparrow$)\end{tabular}} 
& \multicolumn{1}{c}{\begin{tabular}[c]{@{}c@{}}\textbf{Emotion-Aware} \\ \textbf{Diversity} ($\uparrow$)\end{tabular}} 
 \\ 
\hline
&\multicolumn{5}{|c}{\quad\quad\quad\quad\quad\quad\quad\quad\textbf{Compared with SIM Methods}}\\\cline{2-6}
\multicolumn{1}{c|}{\multirow{4}{*}{\begin{tabular}[c]{@{}c@{}}\rotatebox{90}{\textbf{EmoAgent} v.s.}\end{tabular}}} 
& IP2P~\cite{brooks2023instructpix2pix} & \textbf{91.85\%} v.s. 8.15\% & \textbf{85.39\%} v.s. 14.61\% & \textbf{89.63\%} v.s. 10.37\% & \textbf{91.85\%} v.s. 8.15\%  \\ 
\multicolumn{1}{c|}{} & MagicBrush~\cite{zhang2024magicbrush} & \textbf{91.54\%} v.s. 8.46\% & \textbf{87.69\%} v.s. 12.31\% & \textbf{93.33\%} v.s. 6.67\% & \textbf{88.15\%} v.s. 11.85\%  \\\cline{2-6}
&\multicolumn{5}{|c}{\quad\quad\quad\quad\quad\quad\quad\quad\textbf{Compared with  AIM Methods}}\\\cline{2-6}
\multicolumn{1}{c|}{} & CLVA~\cite{fu2022language} & \textbf{92.31\%} v.s. 7.69\% & \textbf{80.77\%} v.s. 19.23\%  & \textbf{91.11\%} v.s. 8.89\% & \textbf{94.08\%} v.s. 5.92\%  \\ 
\multicolumn{1}{c|}{} &  AIF~\cite{weng2023affective} & \textbf{94.62\%} v.s. 5.38\% & \textbf{81.54\%} v.s. 18.46\% & \textbf{88.89\%} v.s. 11.11\% & \textbf{88.15\%} v.s. 11.85\%  \\ 
\multicolumn{1}{c|}{} &  EmoEdit~\cite{yang2024emoedit} & \textbf{80.77\%} v.s. 19.23\% & \textbf{76.92\%} v.s. 23.08\% & \textbf{76.30\%} v.s. 23.70\% & \textbf{80.74\%} v.s. 19.26\% \\ \bottomrule
\end{tabular}}
\vspace{-5mm}
\end{table*}

\begin{figure*}[!t]
    \centering
    \vspace{-5mm}
    \includegraphics[width=\linewidth]{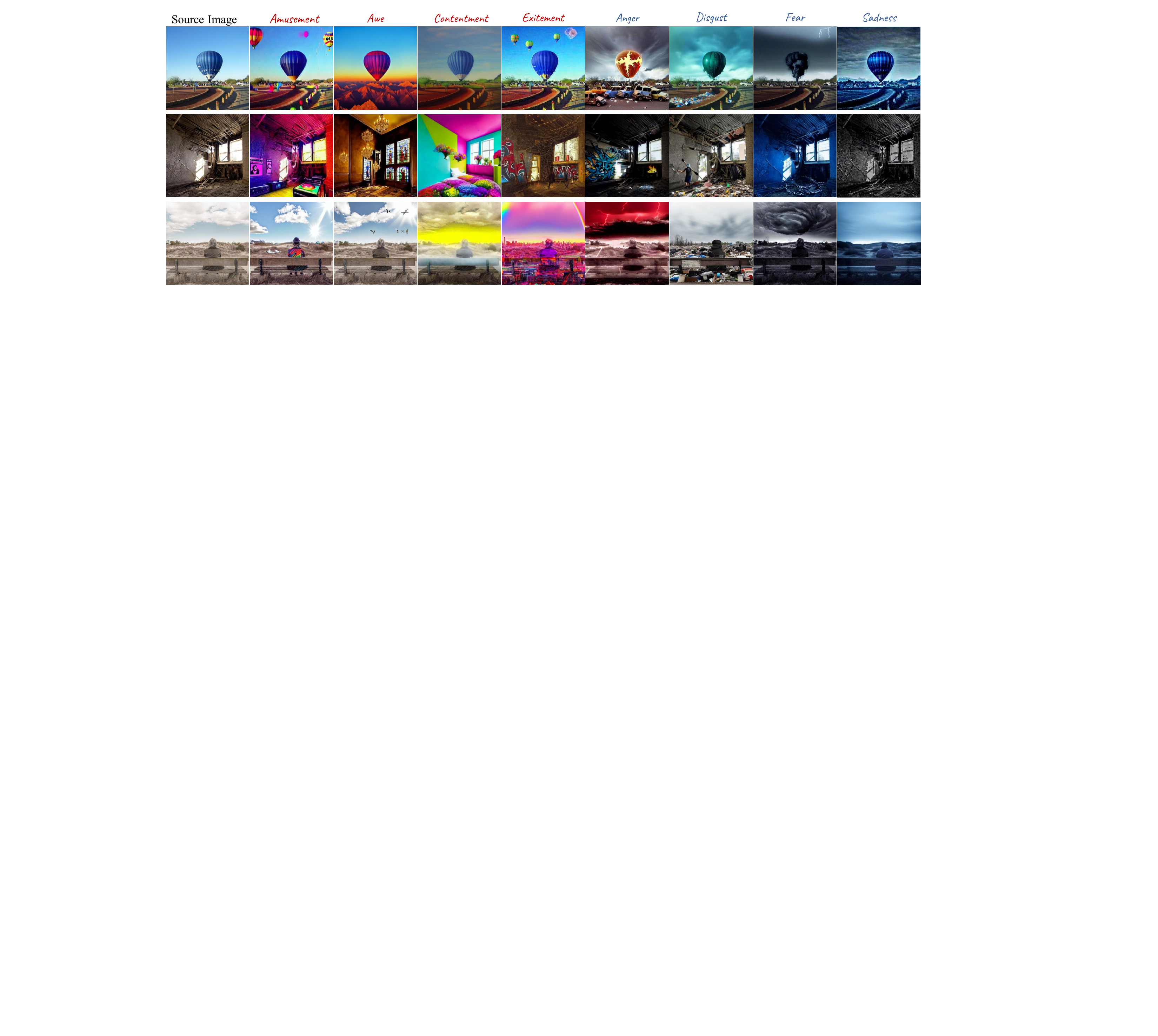}
    \vspace{-7mm}
    \caption{
        \textbf{Results Across Multiple Emotion Categories.}    
EmoAgent successfully manipulates images to reflect eight distinct emotion categories by introducing varied and emotion-relevant editing elements, demonstrating its flexibility across different affective targets.
    }
    \label{fig:multi emotion}
    \vspace{-5mm}
\end{figure*}

\subsection{User Study}
\label{sec:Userstudy}
Given the inherently subjective nature of AIM, we conduct a human evaluation study to assess four key aspects: \emph{emotion accuracy}, \emph{semantic plausibility}, \emph{structural-affective balance}, and \emph{emotion-aware diversity}.
For the first three aspects, participants are shown a source image, a target emotion, and two edited results—one generated by our method and the other by a randomly selected baseline. 
The two edited images are presented in random order to avoid any positional bias. Participants are then asked to compare the outputs based on the specified criterion.
For the diversity evaluation, we present participants with sets of edited outputs generated under the same emotion category from the same source image. Participants are asked to select which set demonstrates greater visual diversity while maintaining consistency with the target emotion.

The specific questions for each task are as follows:

\begin{compactitem}
\item \textbf{Emotion Accuracy}: Compared to the source image, which edited image more accurately conveys the intended emotional transformation?
\item \textbf{Semantic Plausibility}: Which edited image presents a more natural and semantically reasonable transformation in line with the target emotion?
\item \textbf{Structural-Affective Balance}: Which edited result better balances emotional transformation with preservation of the original structure?
\item \textbf{Emotion-Aware Diversity}:  Which set of edited images provides richer and more diverse visual expression of the same emotion?
\end{compactitem}

We randomly select $30$ source images from the test set, each associated with multiple target emotions, and construct a questionnaire covering the four evaluation aspects. 
%
%
A total of $60$ participants were recruited to complete the evaluation.
As summarized in \tabref{userstudy}, our method achieves significantly higher preference rates than baseline methods across all four dimensions. 
Specifically, compared to EmoEdit~\cite{yang2024emoedit}, the current SOTA in AIM, EmoAgent achieves higher user preference scores: \textbf{80.77\%} for Emotion Accuracy, \textbf{76.92\%} for Semantic Plausibility, \textbf{76.30\%} for Structural-Affective Balance, and \textbf{80.74\%} for Emotion-Aware Diversity.
These results demonstrate that our method not only delivers emotionally accurate and semantically meaningful edits but also excels in balancing structural integrity while offering greater visual diversity.
The consistently lower performance of EmoEdit across all four metrics further underscores EmoAgent’s advantages in both standard AIM and D-AIM tasks.

\begin{figure}[!t]
    \centering
    \includegraphics[width=\linewidth]{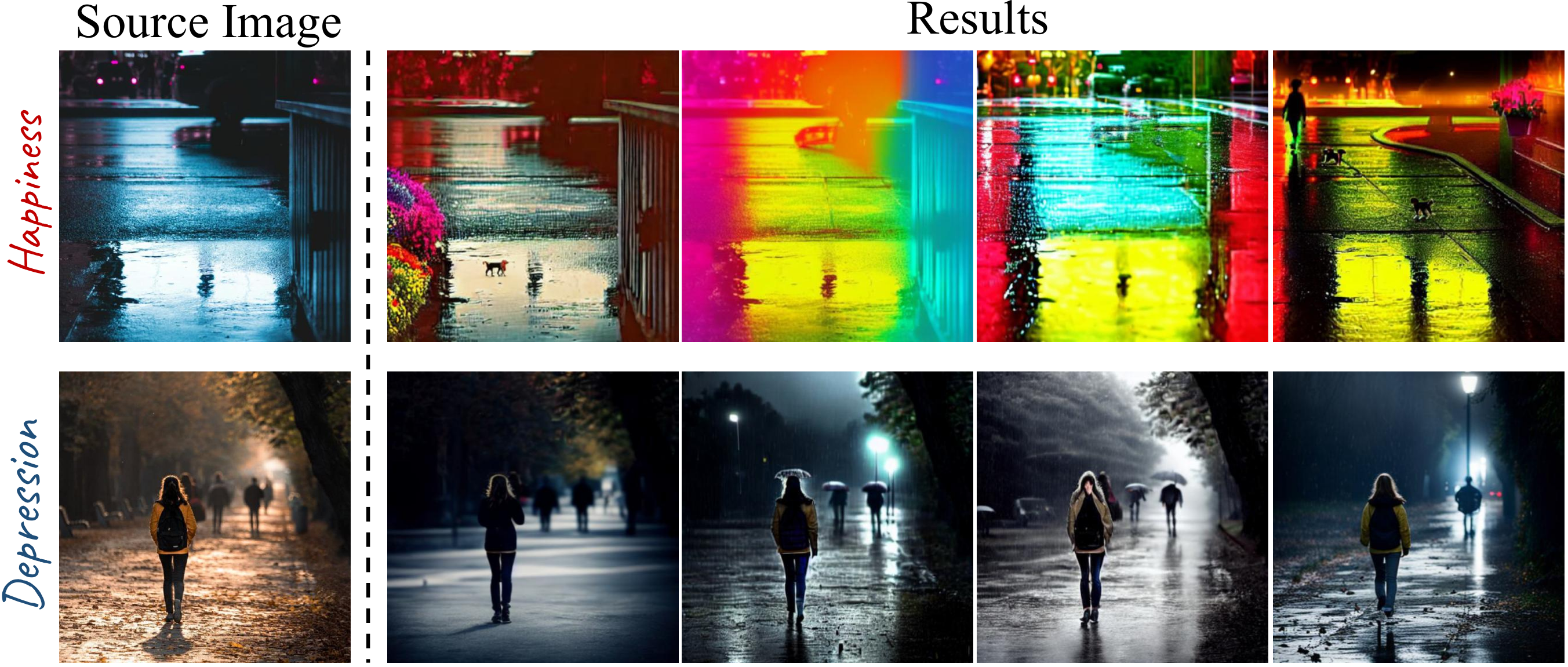} 
    \caption{
            \textbf{Qualitative results on out-of-domain emotions.}
EmoAgent generalizes to novel emotional intents by leveraging the Planning Agent’s in-context reasoning.
    }
    \vspace{-7mm}
    \label{fig:out}
\end{figure}

%% file: sec/5_Discussion.tex
\section{DISCUSSIONS}
\label{sec:DISCUSSION}

\subsection{Manipulation in Various Emotion Categories}
Our EmoAgent is capable of modifying a given image across eight distinct emotional directions, incorporating various semantic factors specific to each emotion category, as illustrated in \figref{multi emotion}.
For instance, as shown in the second row, when presented with an image of a ``dilapidated house'', EmoAgent incorporates elements such as ``posters'' and ``phonograph'', replaces the existing floor material with wooden flooring, and applies color filters to accentuate the emotion of ``amusement''.
Additionally, by altering the ``decor of the house''—such as adding ``chandeliers'' and ``murals'' and incorporating ``golden lighting''—EmoAgent effectively conveys a sense of ``awe''.

\subsection{Generalization to Out-of-Domain Emotions}
 Existing approaches such as EmoEdit~\cite{yang2024emoedit} rely on fixed emotion labels and must be retrained for each new category, leading to inefficient use of training resources.
 In contrast, our EmoAgent supports zero-shot generalization to arbitrary emotions. 
This is enabled by the Planning Agent's in-context understanding of emotional semantics, allowing it to adapt editing strategies to novel emotional targets without additional training.
As illustrated in \figref{out}, we present two examples of out-of-domain emotions.
For \textit{happiness}, EmoAgent introduces elements like flowers, puppies, and warm tones to evoke a joyful atmosphere.
For \textit{depression}, it incorporates night scenes, dark clouds, fallen leaves, and umbrellas to convey a sense of melancholy.
These results demonstrate EmoAgent's ability to adaptively select diverse, context-aware visual elements to express a wide range of emotions beyond predefined categories, broadening the expressive scope of AIM.

%% file: sec/6_Conclusions.tex
\section{Conclusions}
In this work, we introduce \emph{Diverse Affective Image Manipulation (D-AIM)}, a new task that targets the generation of multiple visually distinct yet emotionally consistent edits from a single image and target emotion. 
This addresses a key limitation of prior AIM methods, which typically rely on fixed, one-to-one mappings between emotions and visual features.
To this end, we propose \emph{EmoAgent}, the first multi-agent framework for D-AIM. By dividing the editing process into planning, editing, and critic stages, EmoAgent enables structured exploration of one-to-many emotional transformations, supporting diverse, semantically grounded, and emotionally faithful visual outputs.
Extensive experiments demonstrate that EmoAgent significantly outperforms existing AIM methods in both emotional alignment and visual diversity. 
Looking ahead, we recognize that emotional perception varies across individuals. Future work will explore integrating user-in-the-loop mechanisms and adaptive modeling to personalize affective outputs.

%% file: sec/X_suppl.tex
\section{Summary}

In this supplementary material, we provide detailed implementation details, ablation study results, and additional findings as follows:

\begin{compactitem}
\item we present more implementation details on the benchmark dataset, baselines, automatic evaluation metrics, and human evaluation of EmoAgent in \secref{implementation}.
\item Furthermore, \secref{baselines} provide additional qualitative results of comparison with baseline in AIM task and qualitative results with EmoEdit in Diverse-AIM(D-AIM) task setting to complement the paper.

\item We also include visualizations of the EmoAgent workflow to illustrate how the framework operates and adapts to different emotional contexts in \secref{Visualized}
\end{compactitem}

\subsection{Implementation Details}
\label{sec:implementation}
In this section, we detail the implementation of EmoAgent, covering the construction of the emotion-factor tree, creation of the emotion-factor knowledge database, and deployment of editing tools.

\begin{figure}[!t]
    \centering
    \includegraphics[width=\linewidth]{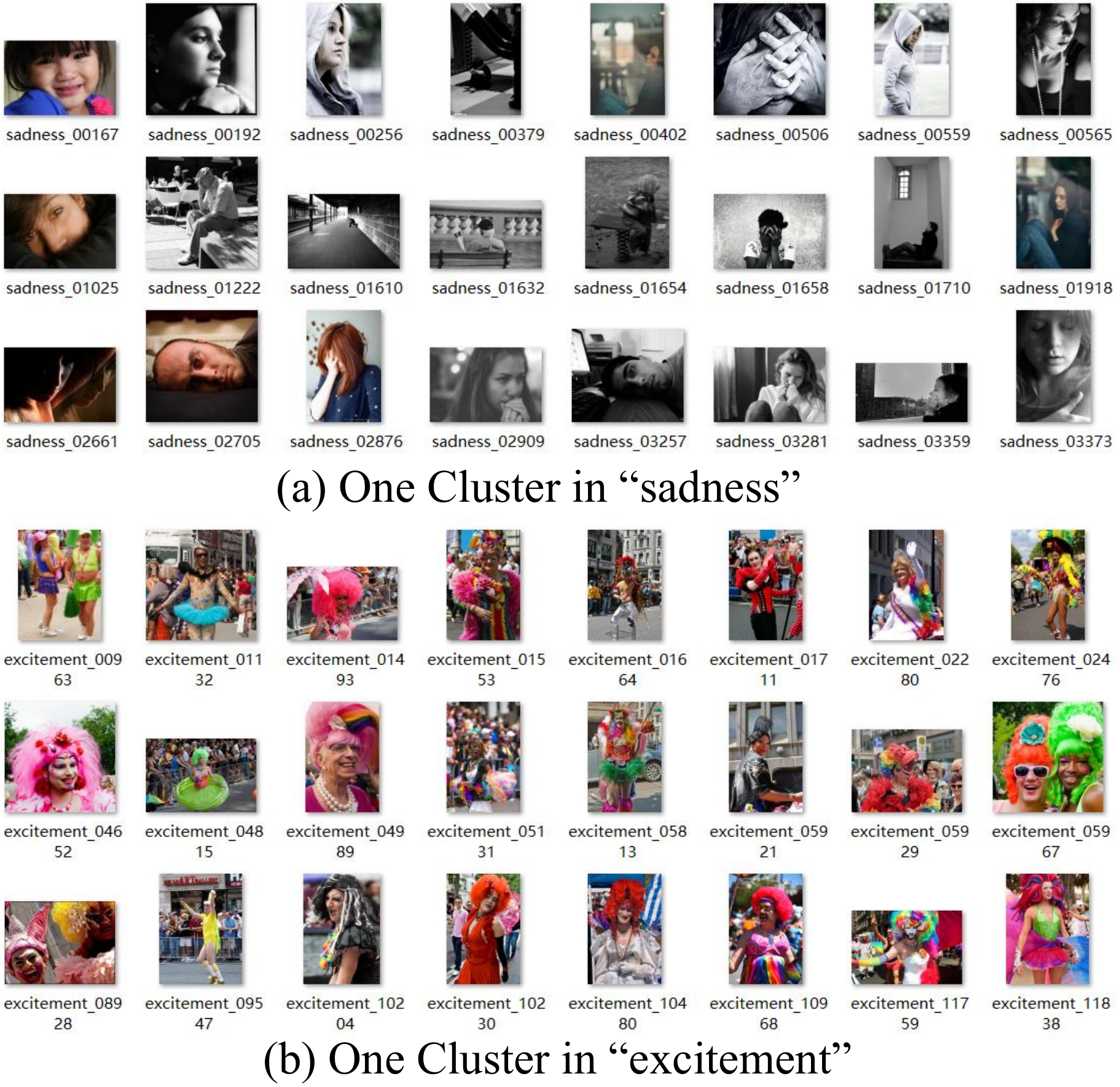}
    \vspace{-6mm}
    \caption{
        \textbf{Cluster examples.}
        (a) An example of cluster results for the ``sadness'' emotion, showing that the main features of this class are people with somber expressions and gray tones.
        (b) An example of cluster results for the emotion of ``excitement'', where the main features of this class are humans with colorful hair and wearing colorful clothes on the street.
    }
    \label{fig:cluster}
    \vspace{-4mm}
\end{figure}

\Paragraph{Details of Emotion-Factor Tree.}
To link subject emotions with specific visual semantic elements, we adopt the strategy of constructing the emotion-factor tree~\cite{yang2024emoedit}. 
The construction process begins by embedding EmoSet~\cite{yang2023emoset} images into a semantic space using the CLIP~\cite{radford2021learning} model, capturing key visual and semantic features for clustering analysis. 
Semantically similar images are grouped together through an iterative clustering approach to identify visual cues associated with distinct emotional categories.
Initially, each image is treated as an independent cluster. 
The two most similar clusters are merged in each iteration until the similarity between all clusters drops below a predefined threshold (\eg 0.89).
To enhance the relevance and generalizability of emotional expressions, we implement a multi-step post-processing procedure that includes: 1) removing clusters with fewer than five images for statistical reliability, 2) excluding clusters with excessive pixel-level similarity to avoid redundancy, and 3) eliminating clusters that do not evoke specific emotions to maintain emotional relevance. 
 The refined clusters, depicted in \figref{cluster}, are then organized into a hierarchical structure based on dimensions such as objects, scenes, actions, and facial expressions. 
Using the GPT-4o~\cite{gpt4o} model, we generate natural language descriptions for each cluster, extracting and summarizing consistent key information across dimensions like ``objects'', ``scenes'', ``facial expressions'', ``actions'', and ``color tones'' to form the hierarchical emotion-factor tree. 
The tree's root nodes represent emotional categories, such as ``sadness'', while leaf nodes detail specific visual elements. 
The extracted semantic information—covering objects, scenes, actions, facial expressions, and colors—is stored in independent JSON files. 

\begin{figure*}[!t]
    \centering
    \includegraphics[width=.75\linewidth]{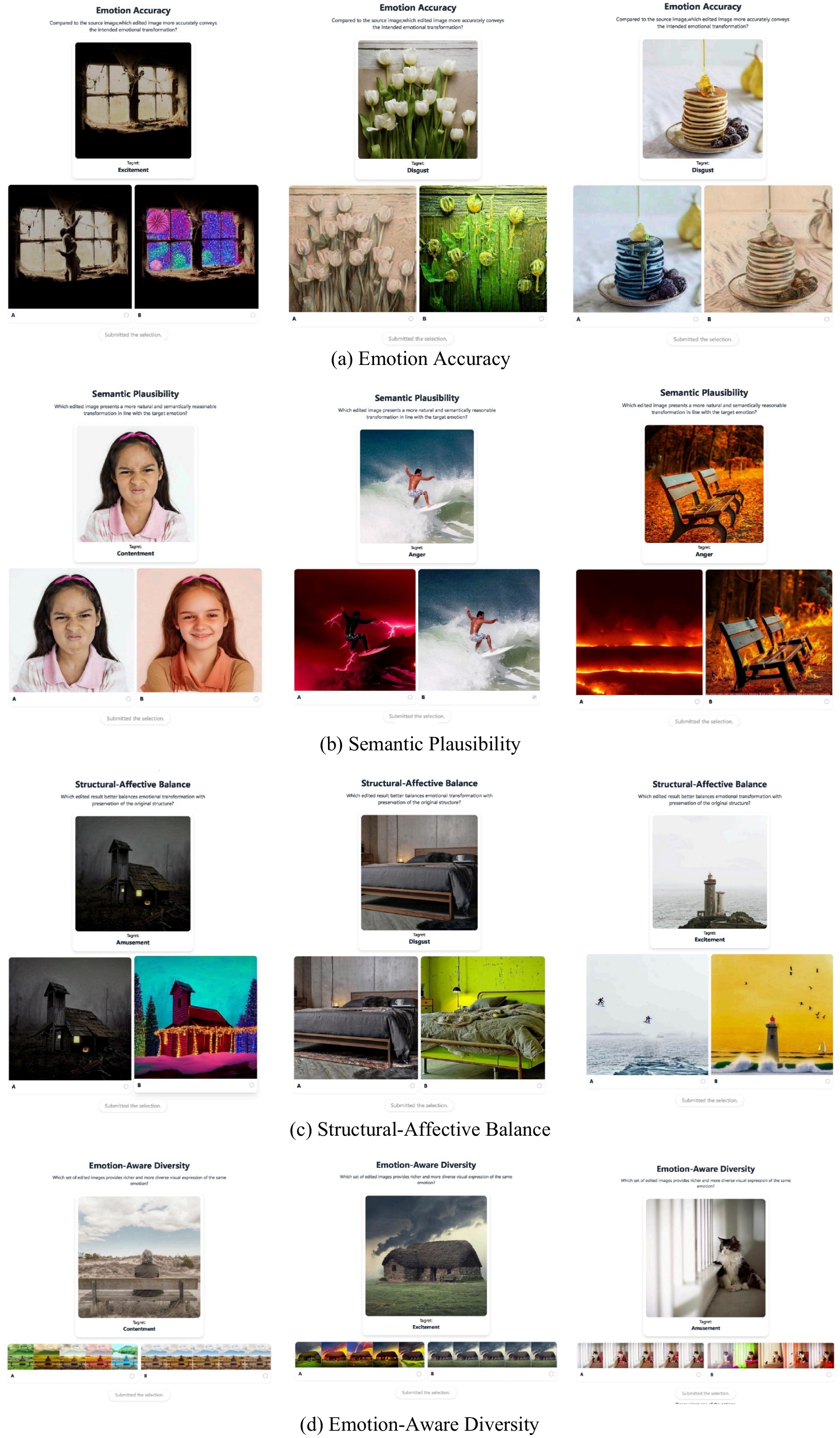}
    \vspace{-3mm}
    \caption{
        \textbf{Examples of tasks for human raters on user study to complete.}
        (a) Exhibition of emotion accuracy questionnaire interface.
        (b) Exhibition of semantic plausibility questionnaire interface.
        (c) Exhibition of structural-affective balance questionnaire interface.
        (d) Exhibition of emotion-aware diversity questionnaire interface.
    }
    \label{fig:userstudy}
    \vspace{-2mm}
\end{figure*}

\Paragraph{Details of Emotion-Factor Knowledge Database.}
The EFK Database performs retrieval by leveraging the node information provided by the emotion-factor tree. In this database, each emotional factor is transformed into a TextNode, with sub-tags for emotion and element type to refine the search process. 
The text description of each emotional factor serves as the content of these nodes. 
To identify the editing element that best aligns with the original image's semantics, we directly compute the $L_2$ similarity between the node content $E_n$ and the image's semantic text extraction embedding $E_s$. 
\begin{equation}
  d = \sqrt{\sum_{i=1}^{n} (E_n - E_s)^2}. 
  \label{eq:important}
\end{equation}
This approach allows for precise matching and enhances the accuracy of emotional expression in image editing.

\Paragraph{Details for Editing Tools.}
We use the official codes released by the authors for implementing IP2P~\cite{brooks2023instructpix2pix}, 
MagicBrush~\cite{zhang2024magicbrush}, PnP~\cite{tumanyan2023plug}, MAG-Edit~\cite{mao2024mag}, Guide~\cite{titov2024guide}, SDXL~\cite{podell2023sdxl}, RF-Solver-Edit~\cite{wang2024taming}, Mimicbrush~\cite{chen2024mimicbrush}, Grounding Dino~\cite{liu2023grounding}, Segmentation Segment Anything~\cite{kirillov2023segment} as editing tools and auxiliary tools.
Specifically, for SDXL, we select the official stable-diffusion-xl-$1.0$-inpainting-$0.1$ model as the editing tool, SDXL-Inpainting, for object replacement and background editing.
Additionally, we employ the official Refiner model, SDXL-Refiner, as an auxiliary tool to optimize the fine-grained details of images.
\Paragraph{Implementation Details of Baselines.}
We employ the official codes released by the authors for implementing IP2P~\cite{brooks2023instructpix2pix}, MagicBrush~\cite{zhang2024magicbrush},  CLVA~\cite{fu2022language}, AIF~\cite{weng2023affective}, EmoEdit~\cite{yang2024emoedit}.
Each algorithm uses information from two modalities as input: the original image and the associated text containing the target emotion.
All algorithms are deployed on the same NVIDIA GeForce RTX $4090$ GPU for consistent performance evaluation.

\Paragraph{Details of Human Evaluation.}
\figref{userstudy}  provides a comprehensive overview of our questionnaire design, which is used for subjective evaluation of emotional image editing results. 
The questionnaire is structured around four key aspects: Emotion Accuracy, Semantic Plausibility, Structural-Affective Balance, and Emotion-Aware Diversity. 
For each aspect, we offer clear definitions, real examples, and detailed descriptions to guide annotators in making consistent and informed judgments. 
These materials ensure transparency and reproducibility in our human evaluation process.

\begin{figure*}[!t]
    \centering
    \includegraphics[width=.9\linewidth]{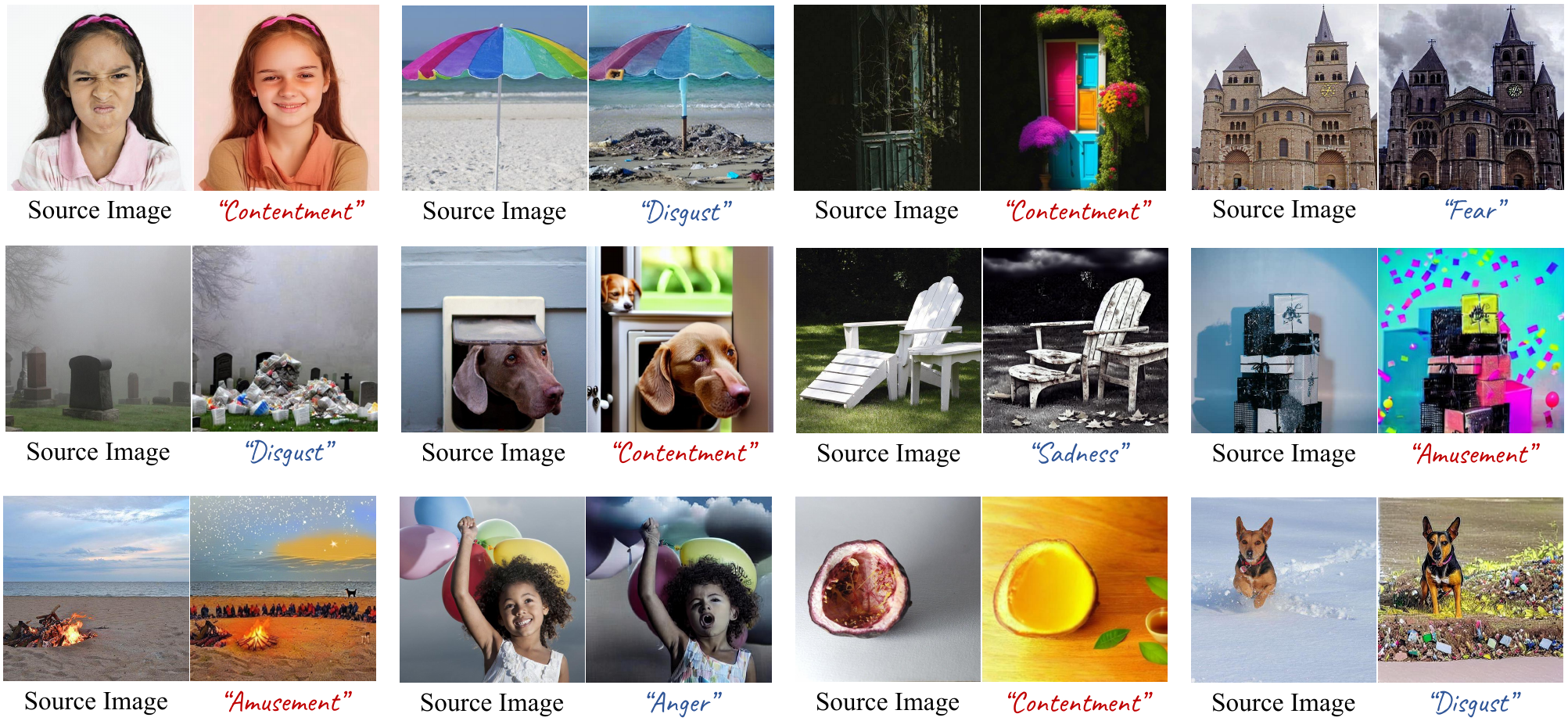}
    \vspace{-3mm}
    \caption{
        \textbf{More subjective experimental results of different emotions.}
         The effectiveness of our EmoAgent in the AIM task can be demonstrated significantly.
    }
    \label{fig:moreresult}
    \vspace{-7mm}
\end{figure*}
\begin{figure*}[!t]
    \centering
    \includegraphics[width=\linewidth]{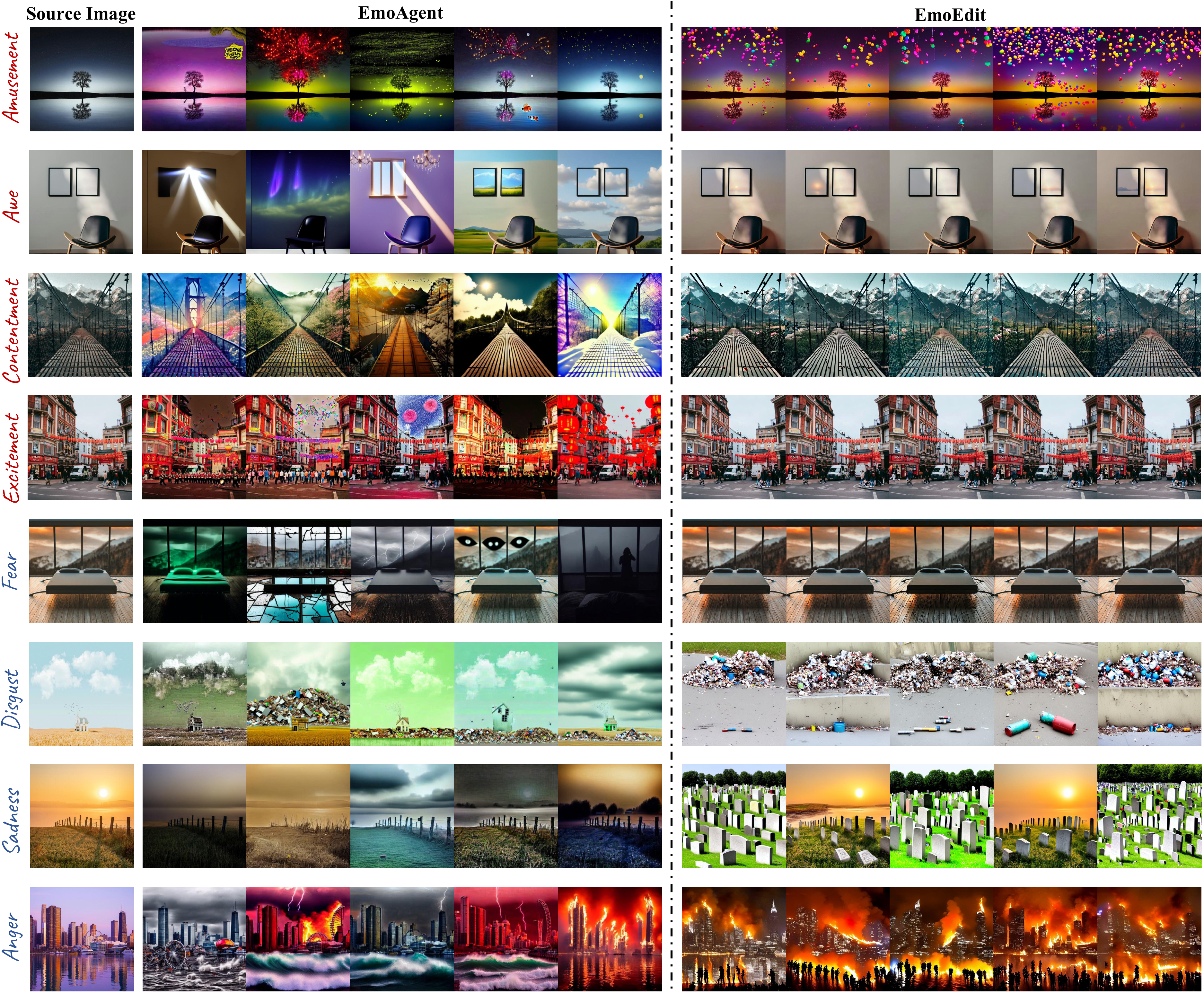}
    \vspace{-6mm}
    \caption{
        \textbf{Additional emotion-aware diversity qualitative results of EmoAgent.}
    }
    \label{fig:more diverse}
    \vspace{-3mm}
\end{figure*}
\subsection{Additional Experimental Results}
\label{sec:baselines}

\begin{figure*}[!t]
    \centering
    \includegraphics[width=.85\linewidth]{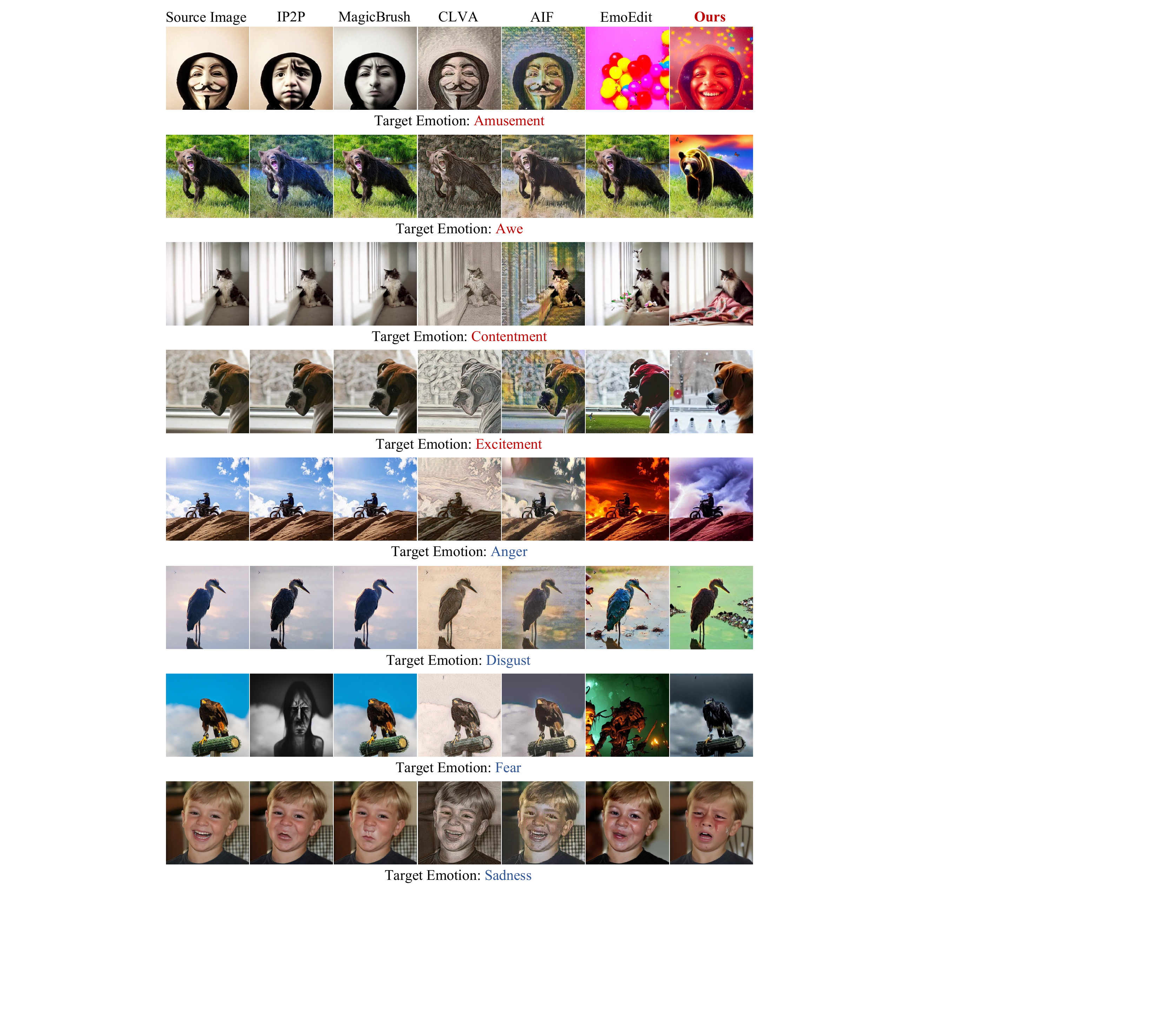}
    \vspace{-4mm}
    \caption{
        \textbf{More qualitative comparisons with existing state-of-the-art SIM and AIM methods.}
         Our EmoAgent demonstrates superior ability in expressing target emotion while preserving the integrity of image structure.
    }
    \label{fig:moreresults}
    \vspace{-5mm}
\end{figure*}

\Paragraph{Additional Qualitative Results.}
We present further qualitative comparisons with existing approaches, as illustrated in \figref{moreresults}. Additional examples of affective image manipulation are illustrated in \figref{moreresult}.
These extensive experimental results underscore the superior performance of our method in the AIM task, demonstrating its effectiveness in accurately conveying intended emotions through targeted image modifications.

\begin{figure*}[!t]
    \centering
    \includegraphics[width=.85\linewidth]{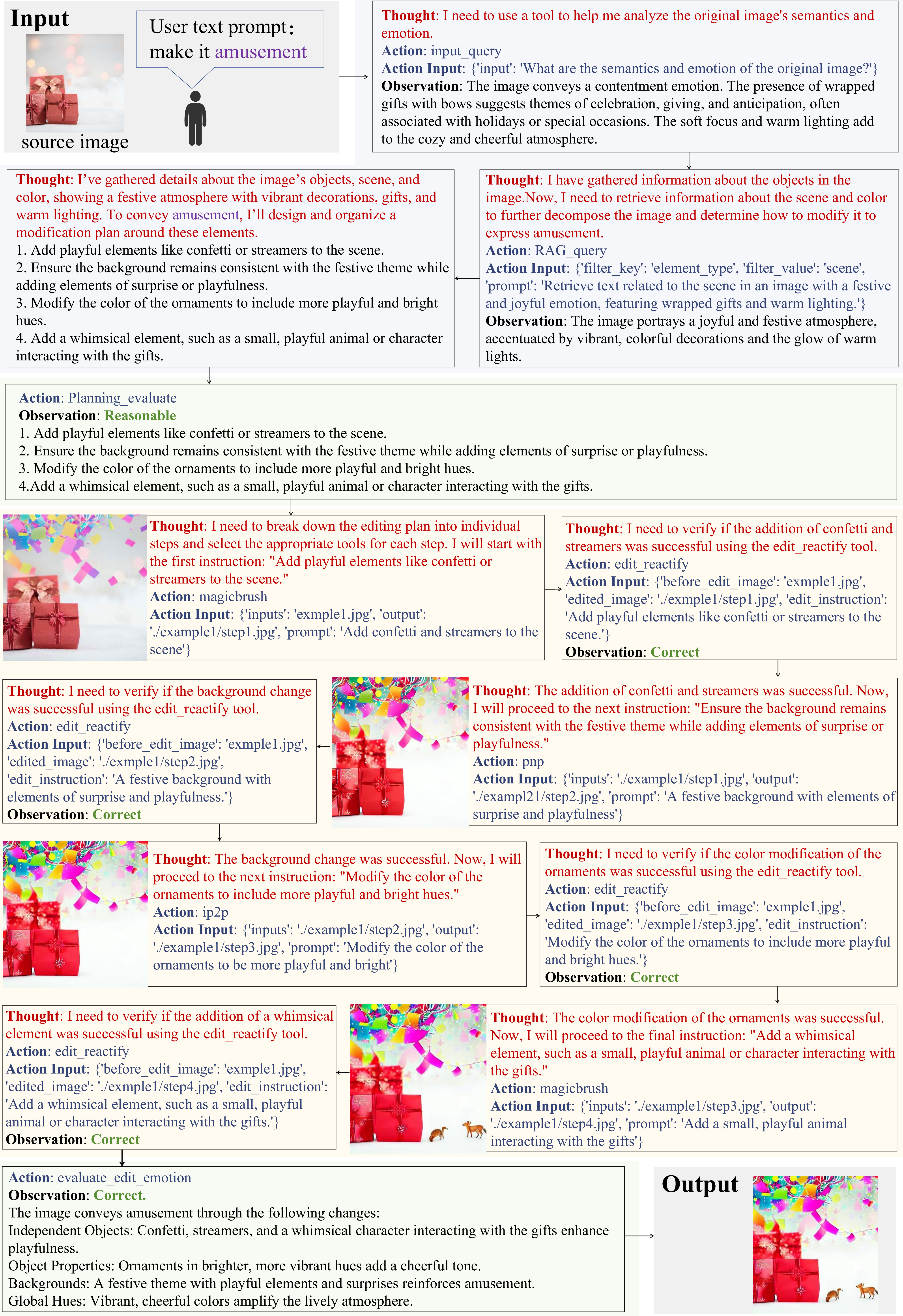}
    \vspace{-4mm}
    \caption{
        \textbf{The first workflow example of EmoAgent.}
        An ideal workflow without any editing or emotional errors.
        The \colorbox{mycolor_bluelight}{blue}, \colorbox{mycolor_green}{green}, and \colorbox{mycolor_yellow}{yellow} process boxes represent the workflows of the Planning Agent, the Critic Agent, and the Editing Agent, respectively.
    }
    \label{fig:workflow}
    \vspace{-7mm}
\end{figure*}
\begin{figure*}[!t]
    \centering
    \includegraphics[width=.845\linewidth]{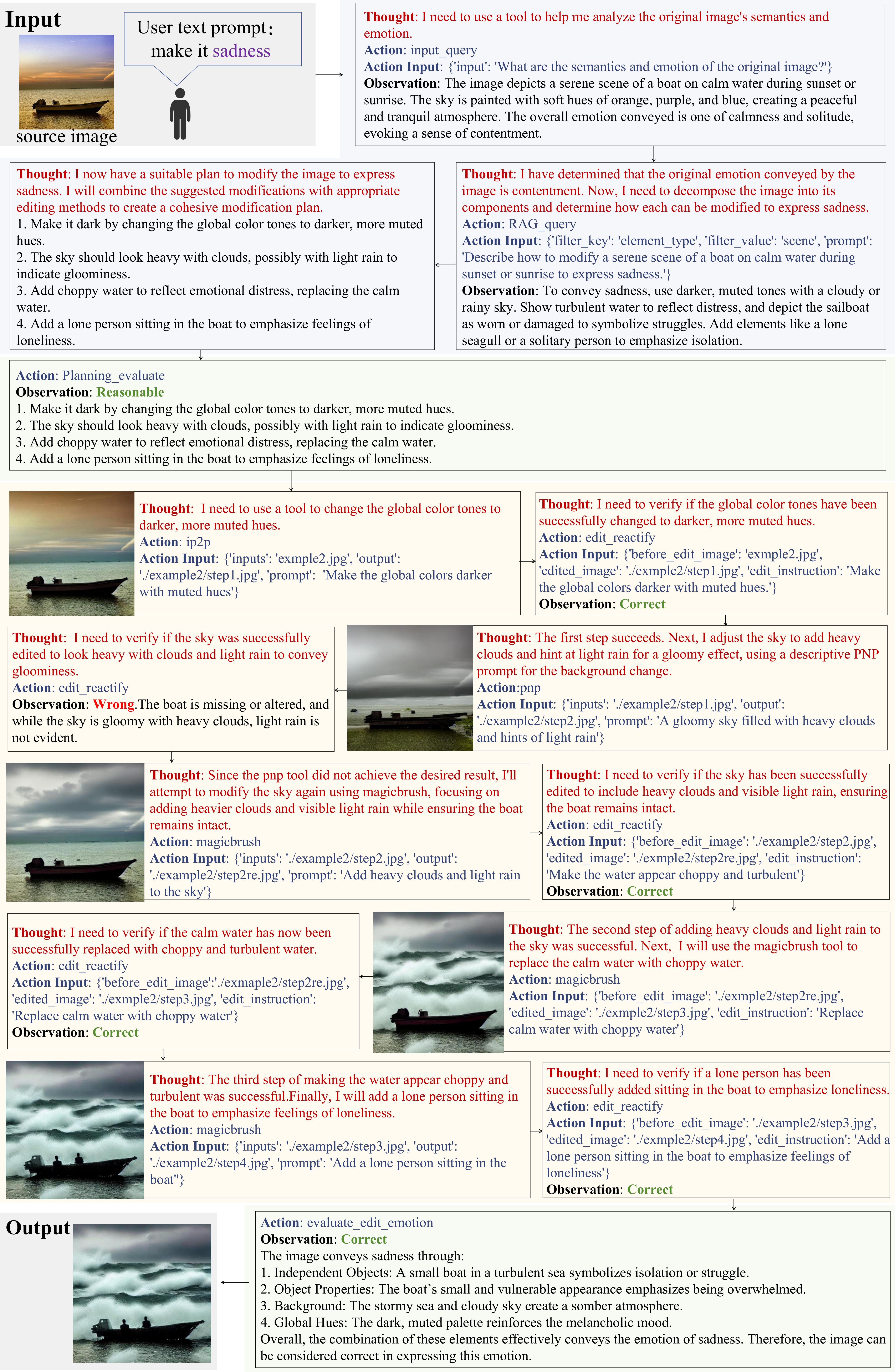}
    \vspace{-4mm}
    \caption{
        \textbf{The second workflow example of EmoAgent.}
        The workflow when EmoAgent encounters editing errors.
        The \colorbox{mycolor_bluelight}{blue}, \colorbox{mycolor_green}{green}, and \colorbox{mycolor_yellow}{yellow} process boxes represent the workflows of the Planning Agent, the Critic Agent, and the Editing Agent, respectively.
    }
    \label{fig:workflow2}
    \vspace{-7mm}
\end{figure*}

\Paragraph{Additional D-AIM Results.}
\label{sec:diverse_more}
In this part, we present additional emotion-aware diversity qualitative results in \figref{more diverse} to demonstrate the performance of our approach on the Diverse-AIM(D-AIM) task setting.

\subsection{Workflow Visualization of EmoAgent}
\label{sec:Visualized}

We provide the visualization examples to demonstrate the concrete editing process of our EmoAgent.
A key advantage of our EmoAgent over other approaches is \emph{its high degree of interpretability at every step of the workflow}. 
This level of transparency enables users to clearly understand and trace the decision-making and emotion processing throughout the system.
In \figref{workflow}, we present a prime example of all processes being executed correctly. EmoAgent adheres to the workflow, efficiently executing each step in sequence.
In \figref{workflow2}, we present a case where the initial editing actions are incorrect, yet are subsequently corrected by the Self-Critic mechanism. 
This process continues until all steps are completed correctly, demonstrating our method's capability for precise image manipulation. 
As shown in \figref{workflow} and \figref{workflow2}, all three agents operate with the working logic of ReAct~\cite{yao2022react} agent, where each action strictly adheres to the workflow of thinking, discovering, and executing.